%% file: main.tex
\documentclass{article}

\PassOptionsToPackage{numbers,sort&compress}{natbib}

 \usepackage[eandd, final]{neurips_2026}

\usepackage[utf8]{inputenc} %
\usepackage[T1]{fontenc}    %
\usepackage{hyperref}       %
\usepackage{url}            %
\usepackage{booktabs}       %
\usepackage{amsmath}        %
\usepackage{amsfonts}       %
\usepackage{nicefrac}       %
\usepackage{microtype}      %
\usepackage{xcolor}         %
\usepackage{graphicx}
\usepackage{rotating}
\usepackage{comment}
\usepackage{booktabs,graphicx,amssymb}
\usepackage{wrapfig}

\usepackage{subcaption}

\newcommand{\martine}{\textsc{Martine}}
\newcommand{\obj}[1]{\textsc{#1}} %

\title{\martine: Benchmarking Multi-View 3D Surface Reconstruction Across Viewpoint Coverage, Resolution, and Lighting}

\author{%
  Robin Bruneau$^{1}$\thanks{Equal contribution.} \quad
  Baptiste Brument$^{2}$\footnotemark[1] \quad
  Frederic Giraud$^{3}$ \quad
  Bastian Sigrist$^{3}$ \\
  \textbf{Sascha Jecklin$^{4}$ \quad
  Philipp Fürnstahl$^{3}$ \quad
  Bjoern Menze$^{1}$ \quad
  Lilian Calvet$^{1,3}$} \\[0.5em]
  $^{1}$University of Zurich, Switzerland \quad
  $^{2}$IRIT, Université de Toulouse, CNRS, Toulouse, France \\
  $^{3}$Balgrist University Hospital, University of Zurich, Switzerland \quad
  $^{4}$Independent researcher
}

\begin{document}

\maketitle

\begin{abstract}
Controlled acquisition setups are widely used for high-accuracy multi-view 3D surface reconstruction, as they allow sensor configuration, image resolution, viewpoint distribution, and illumination to be specified.
Yet benchmarking remains difficult, since reconstruction accuracy depends jointly on the reconstruction paradigm, object properties, and acquisition regime.
Existing benchmarks often restrict viewpoint coverage, resolution, or illumination, which can bias comparisons towards methods suited to the chosen setting.
Others lack accurate 3D ground truth or quantified image-to-geometry alignment, limiting their use for metric surface evaluation. We introduce \martine{}, a benchmark for controlled 3D object surface reconstruction.
\martine{} comprises 28 objects, including transparent materials and fine structures, captured at \(9568 \times 6376\) pixels from 84 views over a spherical cap under both single-light and multi-light regimes.
It provides object-independent camera calibration, accurate 3D ground-truth scans, calibration and image-to-geometry accuracy quantified on test images, and 3D property masks for analysing errors by material properties, visibility, curvature and fine structures.
Experiments across photogrammetry, classical multi-view stereo, neural implicit surfaces, Gaussian-splatting-based reconstruction, and multi-view photometric stereo show that \martine{} enables regime-aware evaluation with explicit control over reference-data accuracy.
The benchmark resolves sub-millimetric reconstruction differences, with a best observed Chamfer distance of \(0.058\) mm over the evaluated visible surface.
\textbf{Project page:} \url{https://martine-benchmark.github.io/}
\end{abstract}

\input{sec/1_introduction}
\input{sec/3_dataset}

\input{sec/4_experiments}

\input{sec/5_conclusion}

\begin{ack}
Robin Bruneau was funded by the SNSF SINERGIA (F01-10266) and the Helmut Horten Foundation. Baptiste Brument was funded by the OPEN-DOPAMIn project (CNRS Innovation). This work has been supported by the OR-X---a Swiss national research infrastructure for translational surgery---and associated funding by the University of Zurich and University Hospital Balgrist. This work was granted access to the Juliet computing platform, supported by a French government grant managed by the Agence Nationale de la Recherche under the ``Investissements d'avenir'' programme (reference ``ANR-21-ESRE-0051''). The authors declare no competing interests.
\end{ack}

\bibliographystyle{unsrtnat}
\bibliography{references_fixed}

\newpage
\input{sec/6_supplementary}

\end{document}

%% file: sec/1_introduction.tex
\section{Introduction}
\label{sec:intro}

3D surface reconstruction is a central problem in computer vision, with applications in cultural heritage preservation, industrial inspection, robotics, augmented reality, and digital content creation. Across these applications, the objective is often not only to recover a visually plausible shape, but to estimate surface geometry with sufficient accuracy for measurement, analysis, or downstream processing. Multi-view RGB surface reconstruction has achieved increasingly accurate results, in particular through advances in photogrammetry software and classical multi-view stereo (MVS)~\cite{metashape,realitycapture,meshroom,colmap,gomvs}, neural implicit surface methods~\cite{neus,na}, and recent multi-view multi-light variants~\cite{supernormal,rnbneus}. While 3D reconstruction can be performed in unconstrained environments~\cite{schops2017eth3d,knapitsch2017tanks,yao2020blendedmvs,reizenstein2021common,jampani2023navi}, controlled acquisition setups provide important advantages when high accuracy is required. Such setups allow the acquisition conditions to be specified in terms of sensor selection and configuration, image resolution, viewpoint distribution and illumination. 

Existing controlled benchmarks cover this acquisition space only partially. DTU~\cite{dtu} and Skoltech3D~\cite{skoltech3d} provide dense but azimuth-limited coverage, rather than broad object-centric coverage for multi-view surface reconstruction. While both include several illumination conditions, the lighting variations are primarily used to evaluate multi-view single-light methods under different static illuminations, rather than to support multi-view photometric stereo, which requires dense illumination sampling. This limitation is quantitatively studied by~\cite{rnbneus2}, which evaluates the suitability of existing controlled datasets for photometric-stereo-based surface reconstruction.

\begin{table}[t]
\centering
\small
\caption{
Comparison of object-scale datasets for controlled surface reconstruction.
\textbf{360$^\circ$ Az.}, \textbf{Multi-Elev.}, \textbf{Dense Lights}, and \textbf{UHD} indicate full azimuthal coverage, multiple elevation levels, dense controlled illumination, and ultra-high-resolution imagery, respectively.
\textbf{Image--GT Res.} indicates whether object-level residuals between evaluation images and the 3D reference geometry are reported, in addition to camera-calibration or scanner-accuracy statistics.
\textbf{Challenge Eval.} indicates whether official masks or groups are provided to analyse errors by surface or material challenge.
\textit{n.r.} means not reported.
}
\label{tab:dataset_comparison}
\makebox[\linewidth][c]{%
\begin{tabular}{lcccccc}
\toprule
Dataset & 360$^\circ$ Az. & Multi-Elev. & Dense Lights & UHD & Image--GT Res. & Challenge Eval. \\
\midrule
DTU~\cite{dtu}                     & $\times$      & $\checkmark$ & $\times$      & $\times$      & \textit{n.r.} & $\times$ \\
LUCES-MV~\cite{lucesmv}            & $\checkmark$ & $\times$      & $\checkmark$ & $\times$      & \textit{n.r.} & $\times$ \\
DiLiGenT-MV~\cite{diligentmv}      & $\checkmark$ & $\times$      & $\checkmark$ & $\times$      & \textit{n.r.} & $\times$ \\
Skoltech3D~\cite{skoltech3d}       & $\times$      & $\checkmark$ & $\times$      & $\checkmark$ & \textit{n.r.} & $\times$ \\
OpenSubstance~\cite{opensubstance} & $\checkmark$ & $\checkmark$ & $\checkmark$ & $\times$      & \textit{n.r.} & $\times$ \\
EvalMVX~\cite{evalmvx}             & $\checkmark$ & $\times$      & $\checkmark$ & $\times$      & \textit{n.r.} & $\times$ \\
Ours                                & $\checkmark$ & $\checkmark$ & $\checkmark$ & $\checkmark$ & $\checkmark$  & $\checkmark$ \\
\bottomrule
\end{tabular}
}
\end{table}

Conversely, datasets designed for multi-view photometric stereo combine several viewpoints with multiple controlled illumination conditions, so that surface geometry can be inferred not only from multi-view consistency but also from shading variation. These datasets provide illumination conditions suitable for photometric reconstruction, but each covers the remaining benchmark dimensions only partially, such as viewpoint density, elevation coverage, image resolution, object count, or surface ground truth.
DiLiGenT-MV~\cite{diligentmv} remains the standard benchmark for multi-view photometric stereo, but contains only five objects, each captured from 20 views at \(612 \times 512\) resolution. More recent datasets increase resolution or object count, but remain limited in viewpoint elevation coverage: LUCES-MV~\cite{lucesmv} contains 12 views at \(2368 \times 1552\) resolution, and EvalMVX~\cite{evalmvx} contains 20 views at \(2448 \times 2048\) resolution; all three acquire views at a single elevation.
Such design choices can also bias cross-paradigm comparisons. EvalMVX~\cite{evalmvx}, for example, compares single-light and multi-light reconstruction using 20 views over \(360^\circ\) azimuth at a single elevation, rather than a denser single-light regime such as the widely used DTU benchmark~\cite{dtu}, which provides 49--64 views over approximately \(0^\circ\)--\(110^\circ\) azimuth.
OpenSubstance~\cite{opensubstance} substantially improves view and illumination coverage, with 270 viewpoints and more than 1000 lighting conditions, and is a valuable resource for novel-view synthesis, appearance modelling, and object relighting.
\begin{wrapfigure}{r}{0.48\linewidth}
\vspace{-0.6em}
\centering
\includegraphics[width=0.48\linewidth]{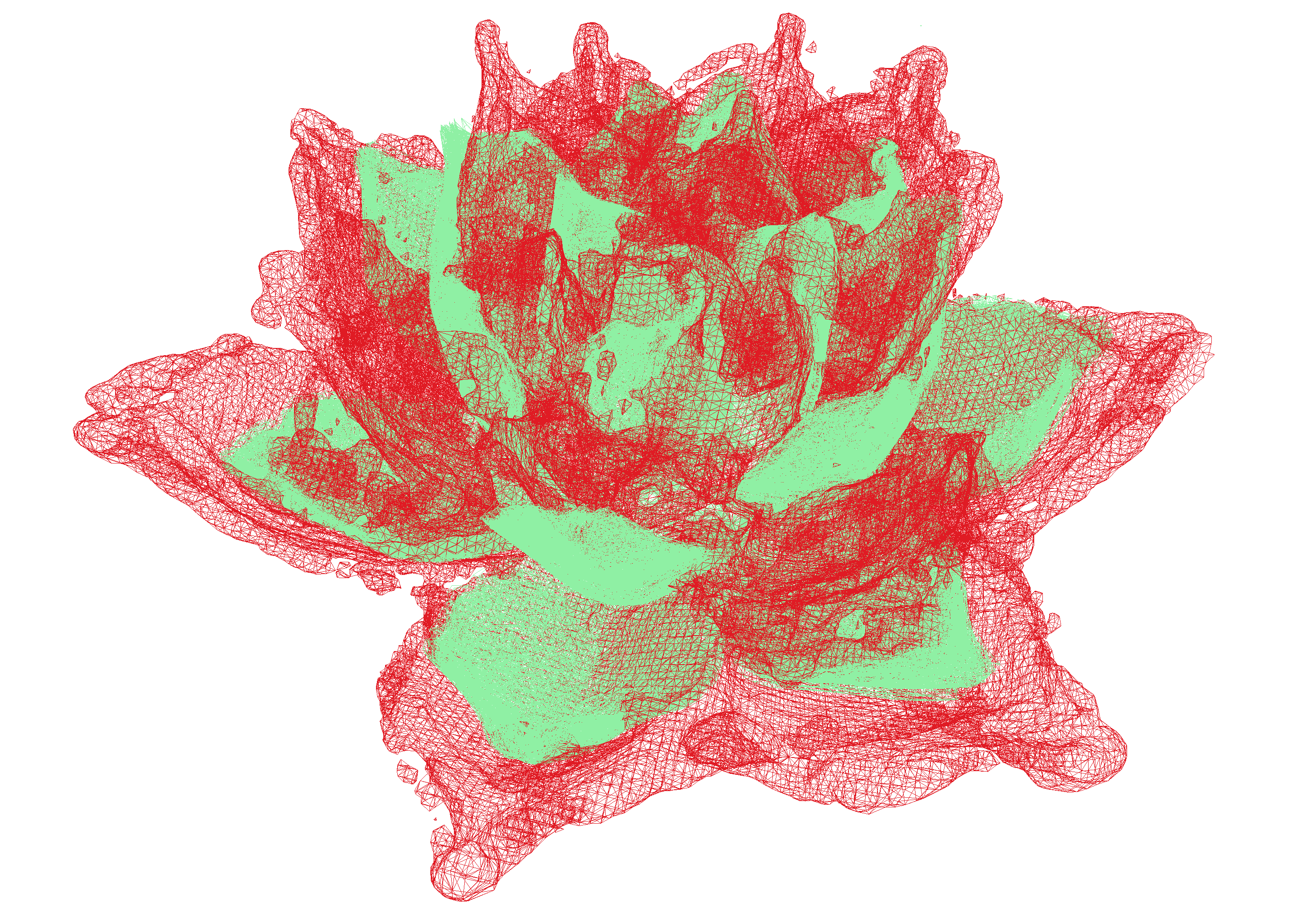}
\vspace{0.3em}
\includegraphics[width=0.48\linewidth]{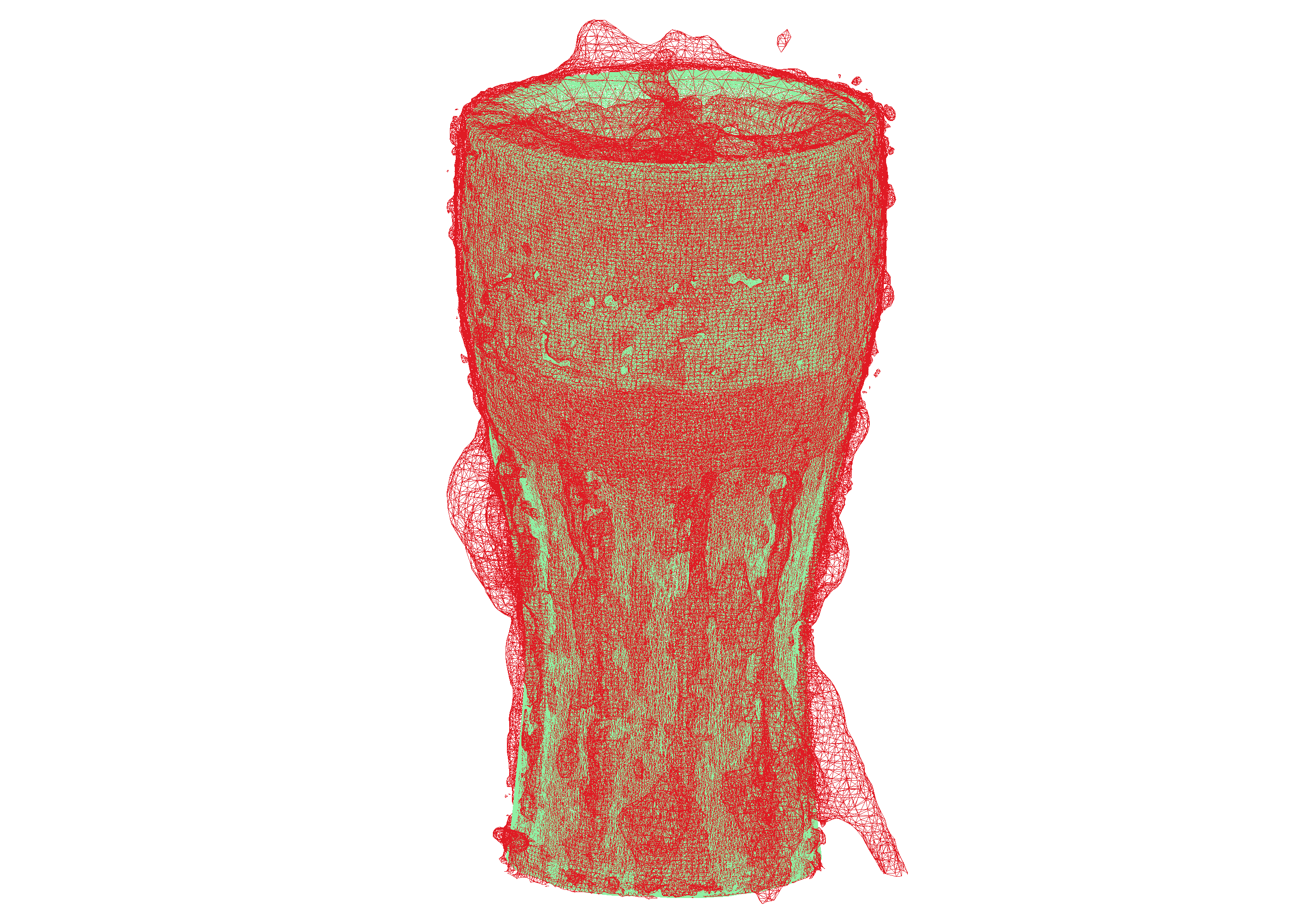}
\caption{
Examples of image--GT misalignment in OpenSubstance~\cite{opensubstance}. Inaccurate COLMAP reconstructions of an opaque flower (left) and a transparent glass (right) in red. Registered GT scan meshes in green.
}
\label{fig:opensubstance_alignment}
\vspace{-1em}
\end{wrapfigure}
However, for surface-reconstruction benchmarking, its image-to-geometry alignment relies on object-dependent registration between the ground-truth scan mesh and a COLMAP~\cite{colmap_mvs} reconstruction. 
This can be unreliable when the COLMAP reconstruction is incomplete or 
geometrically inaccurate (see Figure~\ref{fig:opensubstance_alignment}),
for example for transparent objects, thin structures, or challenging materials;
yet the resulting alignment errors are not systematically quantified or reported.
A related concern is addressed by NAVI~\cite{jampani2023navi} for in-the-wild object reconstruction, where 3D scanner ground truth is 
combined with manually verified 2D--3D alignment
to obtain accurate object alignments.
In addition, OpenSubstance ground-truth scans are not publicly provided for all objects, which limits surface-reconstruction evaluation on challenging cases. 
Similarly, LUCES-MV~\cite{lucesmv} does not provide object-to-camera reference alignment for its transparent objects.

A further issue is that the RGB input used by single-light reconstruction methods in multi-view photometric-stereo datasets is often synthesised from the multi-light stack, for example by taking the median~\cite{kaya2022uncertainty,kaya2023mvpsrevisited} or average~\cite{lucesmv} image over illuminations at each viewpoint. This aggregation has no explicit image-formation basis for single-light reconstruction and is likely motivated by the mitigation of specularities, since specular highlights are often localised and therefore appear only in a subset of the illumination stack. These results remain valuable, but they correspond to a hybrid setting, namely multi-light acquisition followed by single-light reconstruction. They therefore do not directly measure the performance of single-light reconstruction methods under a single-light acquisition regime. As a result, current benchmarks do not jointly provide broad object-centric viewpoint coverage, dense camera-mounted illumination, high-resolution imagery, accurately aligned 3D ground-truth scans, and acquisition regimes that allow single-light and multi-light methods to be compared under conditions designed for each paradigm.

Beyond viewpoint and illumination coverage, image resolution is another source of ambiguity in surface-reconstruction evaluation. This is particularly important for mature photogrammetry pipelines \cite{meshroom,realitycapture}, whose patch-based matching procedures can benefit directly from high-resolution imagery. While higher-resolution images provide more information, they do not consistently lead to lower reported reconstruction errors across existing benchmarks. As reported in~\cite{rnbneus2}, LUCES-MV~\cite{lucesmv} and Skoltech3D~\cite{skoltech3d} yield higher errors than DiLiGenT-MV~\cite{diligentmv}, despite substantially higher image resolutions, \(2368 \times 1552\) and up to \(7296 \times 5472\), compared to \(612 \times 512\). This discrepancy is difficult to interpret because resolution is confounded with dataset-specific differences in acquisition, objects, reference scans, and image-to-geometry alignment. It may indicate either that the reference data are not accurate enough for high-resolution evaluation, or that current implementations do not fully exploit high-resolution inputs. Distinguishing between these explanations requires benchmarks with high-resolution imagery, accurate 3D ground truth and quantified image-to-geometry alignment errors.

A further limitation concerns how reconstruction errors are reported. Most benchmarks report aggregate scores per object \cite{dtu,skoltech3d,diligentmv,evalmvx,opensubstance,lucesmv}, although each object combines several geometric and optical properties. This makes it difficult to isolate failures due to visibility, curvature, fine structures, transparency, specularity, low texture, or other factors. Prior work has already shown that improvements on high-curvature or low-visibility regions can be poorly reflected by global Chamfer-distance or F-score averages~\cite{rnbneus,rnbneus2}. However, such analyses remain scattered across individual papers and are rarely integrated into public benchmark protocols.

To address these limitations, we introduce \martine{}\footnote{Dataset: \url{https://doi.org/10.7910/DVN/2BEPEH}; download and conversion tools: \url{https://github.com/martine-benchmark/martine-dataset}.}, a controlled benchmark for 3D object surface reconstruction. Its main contributions are as follows.

\begin{itemize}
    \item \textbf{A benchmark spanning acquisition regimes.}
    \martine{} comprises 28 objects placed on a motorised turntable and imaged with a high-end camera mounted on a robotic arm. For each object, we acquire 84 views over a spherical cap at \(9568 \times 6376\) resolution under two lighting regimes. The single-light regime uses a reflecting cone and a softbox flash to approximate rotation-invariant ambient illumination around the turntable axis, while the multi-light regime uses 30 camera-mounted LED light sources activated sequentially. This design provides acquisition conditions suited to dense-view single-light reconstruction as well as multi-view multi-light reconstruction. The dataset includes challenging geometries and materials, including fine structures, transparent objects, low-texture surfaces, and specular or glossy regions.
    \item \textbf{Reference data with quantified accuracy.}
    \martine{} provides object-independent camera calibration, calibration accuracy measured on test views, accurate 3D ground-truth scans, and easy-to-interpret 2D--3D registration residuals on test images. These elements make it possible to evaluate surface reconstruction while explicitly accounting for reference-data accuracy.
    \item \textbf{Challenge-level performance analysis.}
    We provide evaluation code and 3D ground-truth masks to report reconstruction errors by optical properties, visibility, curvature and fine structures, rather than only by object identity.
    \item \textbf{A first cross-paradigm evaluation.}
    We provide an initial evaluation of representative methods from photogrammetry, patch-based MVS, 3D Gaussian Splatting-based reconstruction, neural implicit surfaces, and multi-view photometric stereo.
\end{itemize}
Together, these contributions make \martine{} a benchmark for studying how acquisition regime, object properties, and reconstruction paradigm jointly affect surface-reconstruction accuracy.

%% file: sec/3_dataset.tex
\section{Dataset and Acquisition Design}
\label{sec:dataset}

\subsection{Overview}

\begin{figure}[ht!]
    \centering
    \begin{subfigure}[b]{0.555\textwidth}
         \centering
         \includegraphics[width=\textwidth]{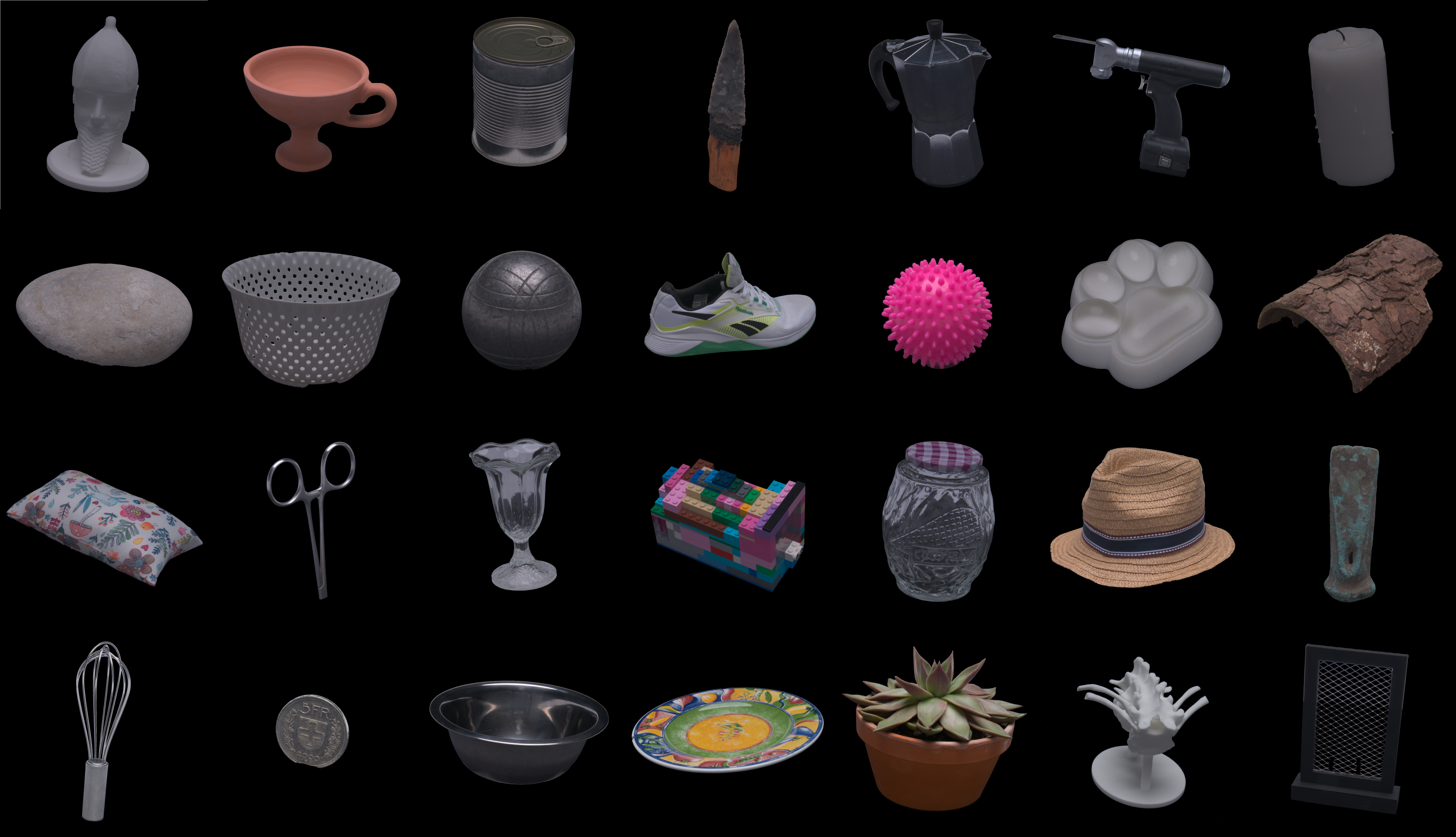}
         \caption{}
         \label{fig:dataset_overview_objects}
     \end{subfigure}
     \hfill
     \begin{subfigure}[b]{0.20\textwidth}
         \centering
         \includegraphics[width=\textwidth]{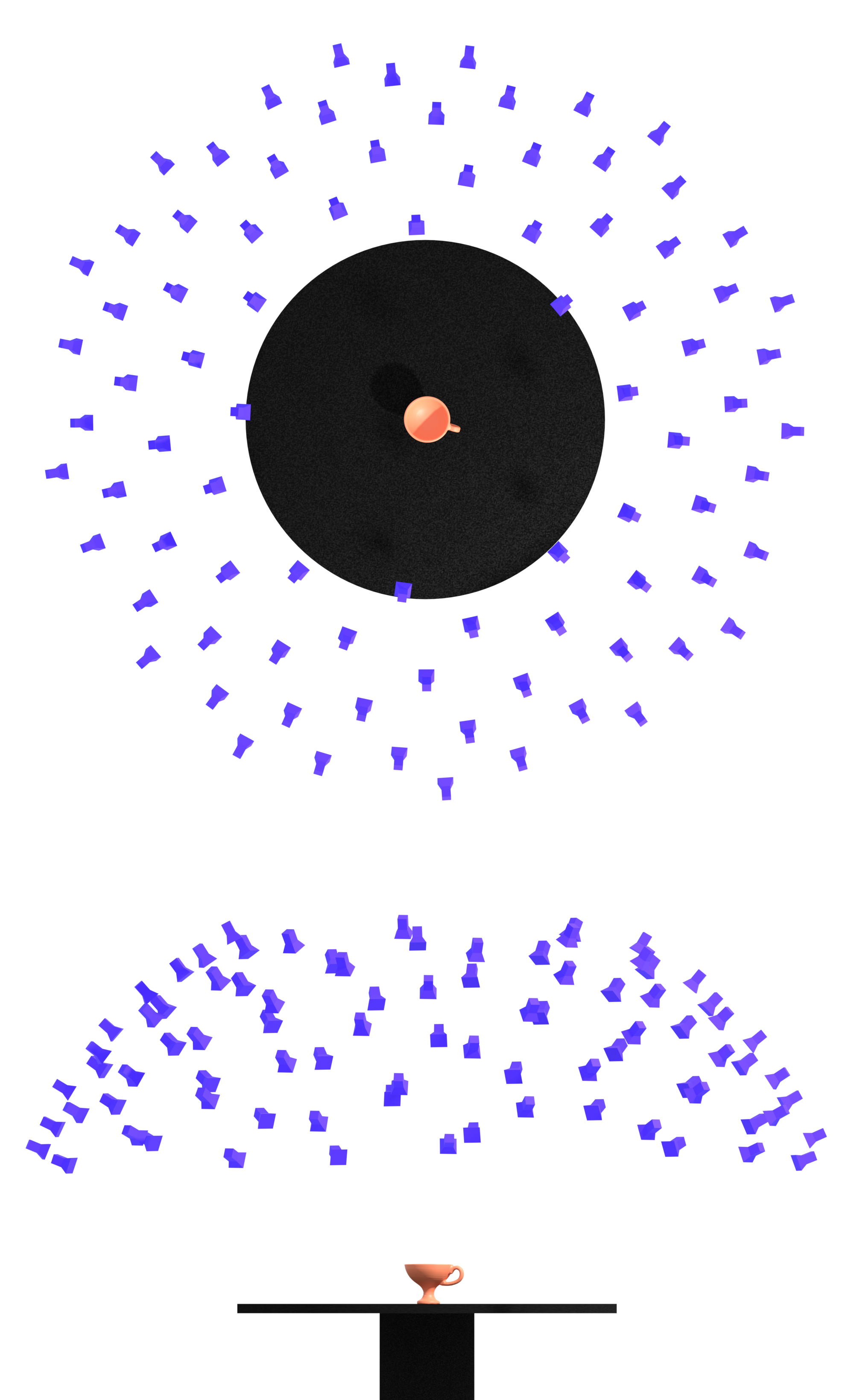}
         \caption{}
         \label{fig:dataset_overview_viewpoints}
     \end{subfigure}
     \hfill
     \begin{subfigure}[b]{0.211\textwidth}
         \centering
         \includegraphics[width=\textwidth]{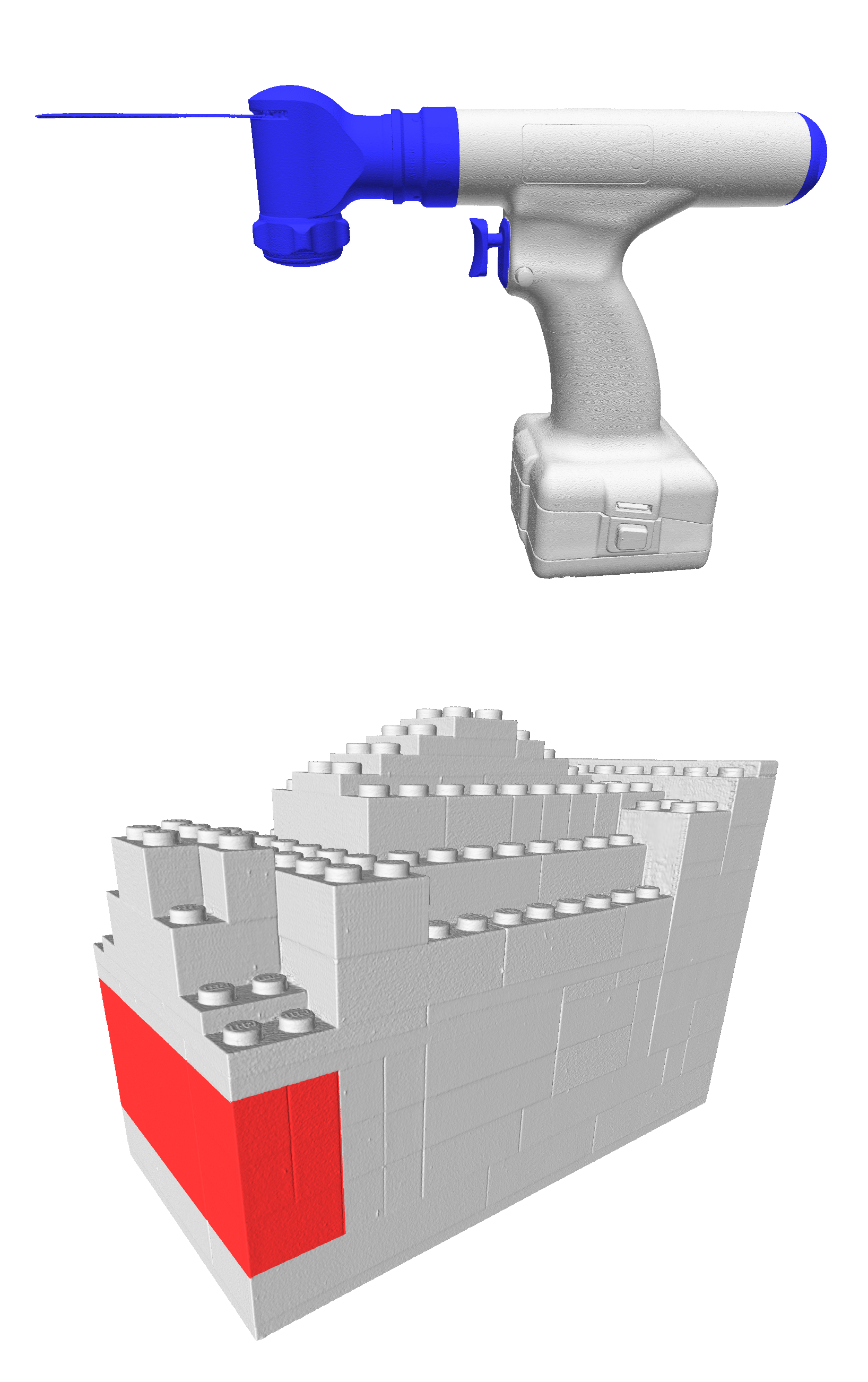}
         \caption{}
         \label{fig:dataset_overview_masks}
     \end{subfigure}
\caption{
Dataset overview.
(a) All objects in the proposed dataset, showing high geometric and optical diversity, including low-texture regions, fine structures, transparent and specular materials, and high-frequency geometric details.
(b) Dense viewpoint sampling with wide angular coverage, obtained by coupling a robotic arm with a turntable.
(c) Two examples of ground-truth meshes with manually curated challenge-level 3D masks highlighting specular regions (top, in blue) and transparent regions (bottom, in red) for challenge-level evaluation.
}
\label{fig:dataset_overview}
\end{figure}

\textbf{Objects.} The object set is designed to cover a range of geometric and optical properties relevant to image-based surface reconstruction. It includes simple textured and matte objects, as well as more challenging cases with fine structures, deep concavities, transparent objects, low-texture surfaces, and highly specular materials. Objects are shown in Figure~\ref{fig:dataset_overview}(a).

\textbf{Viewpoints.} The proposed viewpoints follow a Fibonacci-sphere distribution restricted to a spherical cap of 80 cm radius around the object. We acquire 84 views between \(20^\circ\) and \(70^\circ\) elevation, providing near-uniform sampling of the cap. Each view is captured at \(9568 \times 6376\) resolution with a Sony $\alpha$7R V camera. Viewpoints are shown in Figure~\ref{fig:dataset_overview}(b).

\textbf{Single-light setup.} The single-light regime uses an ambient illumination setup designed to be approximately rotationally symmetric around the turntable axis. This yields nearly view-invariant lighting for the object volume above the turntable, and provides favourable conditions for standard multi-view reconstruction methods. In particular, it substantially reduces the view-dependent appearance changes that would occur if the object were rotated under a fixed asymmetric light source. See Figure~\ref{fig:setup_overview}(a). The rotational symmetry of the illumination was evaluated quantitatively by analysing intensity variations along the isophotes of a white 25 cm sphere placed at the centre of the turntable.
The measured standard deviation across viewpoints ranges from 0.5\% to 2.7\% of the intensity range (mean 1.4\%).
See details in the appendix (Section~\ref{sec:eval_inv_light}).

\textbf{Multi-light setup.} The multi-light regime uses a rig of 30 LEDs rigidly mounted around the camera.
This number of illumination conditions is consistent with the quantitative evaluation of the state-of-the-art photometric stereo method UniMS-PS~\cite{unimsps}, which reports no improvement from increasing the number of input lights beyond 30 in their setting.
It also reflects a practical trade-off in our acquisition design, as each additional light source multiplies the number of images, while the dataset already provides high-resolution 16-bit imagery, dense viewpoints, and two acquisition regimes.
This design provides dense illumination for multi-view multi-light reconstruction. See Figure~\ref{fig:setup_overview}(b).

\textbf{Challenge-level masks.} We provide 3D masks on the ground-truth geometry to support evaluation beyond object-level averages. The masks identify Lambertian or matte regions, textureless regions, transparent parts, specular regions, high-curvature regions and fine structures. The material and structure masks (Lambertian, textureless, transparent, specular, fine structure) are segmented manually on the ground-truth mesh, as conservatively as possible: only regions whose property is unambiguous are labelled, and regions where the label was uncertain are excluded from the evaluation. High curvature is computed automatically from the ground-truth geometry. We also provide visibility levels, defined from the number of views observing each surface point. Low-visibility regions generally correspond to concavities or self-occluded areas, and can also be affected by inter-reflections. Examples of 3D masks are shown in Figure~\ref{fig:dataset_overview}(c).

\subsection{Acquisition Setup}

\begin{figure}[ht!]
    \centering
    \includegraphics[width=\textwidth]{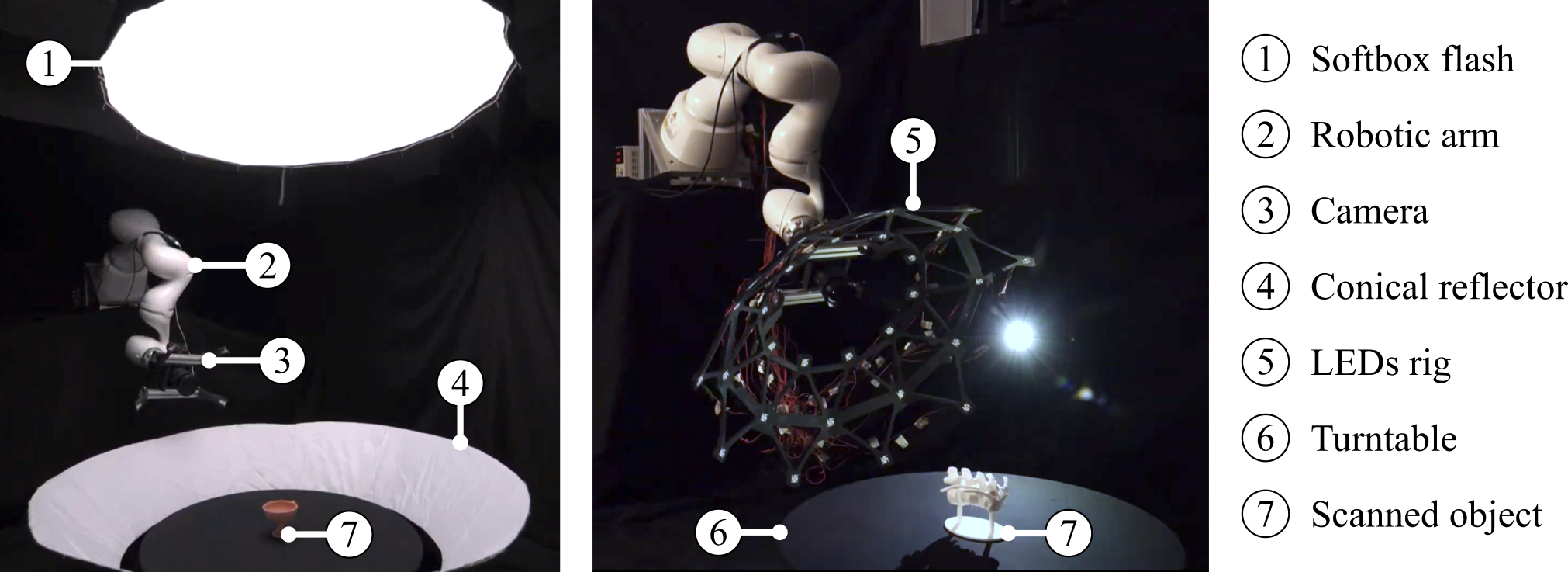}
     
\caption{
Acquisition setup.
(a) Single-light acquisition regime, using an ambient illumination setup designed to be rotation-invariant around the turntable axis.
(b) Multi-light acquisition regime, using a rig of 30 LEDs rigidly mounted around the camera.
At each viewpoint, the LEDs are activated sequentially while the camera pose remains fixed.
}
\label{fig:setup_overview}
\end{figure}

\textbf{View acquisition. } The setup incorporates a Kuka LBR Med 14 industrial robotic arm and a Noxon 360° turntable with an 80 cm diameter. The combination of these two tools allows us to define viewpoints derived from a Fibonacci sphere. This specific trajectory is achieved physically through the ROS 2-managed synchronisation of 10 turntable rotations and precise robotic arm movements (see Figure~\ref{fig:setup_overview}). The turntable is reset between the single-light and multi-light acquisition sequences. This step is intended to maintain consistency between the two capture modes.

\textbf{Lighting setup.}
We acquire each object under two illumination regimes (Figure~\ref{fig:setup_overview}(a) and (b)).
In the single-light regime, the setup uses a 120 cm softbox flash together with a reflecting cone placed around the turntable to produce an approximately rotation-invariant ambient illumination around the object.
In the multi-light regime, illumination is provided by a spherical-cap rig containing 30 Luxeon Rebel LEDs of 220 lm each.
The 85 cm-diameter rig is rigidly mounted to the camera, ensuring a fixed relative configuration between the camera and the light sources across viewpoints.
The LEDs cover a range of lighting directions around the optical axis while remaining at approximately the same distance from the object.
Acquisition is performed in a darkroom, and the LEDs are activated sequentially following a one-light-at-a-time protocol. A black image per viewpoint is also captured for the calibration.

\subsection{Ground Truth Acquisition}

We acquire reference geometry with an Artec Spider optical scanner~\cite{artec}, specified at 0.05 mm 3D point accuracy and 0.1 mm 3D resolution.
As in high-accuracy object-reconstruction benchmarks such as Skoltech3D~\cite{skoltech3d}, specular and transparent objects are sprayed with Aesub Blue, a sublimating opacifying spray, to reduce optical artefacts during scanning.

The scans are acquired manually rather than with a fixed scanner trajectory.
This choice is more time-consuming, but gives the operator direct control over the sensor path and allows coverage to be checked continuously during acquisition.
It is particularly useful for objects with concavities, thin structures, or self-occluded regions, where a fixed scanner and turntable setup may leave parts of the surface insufficiently observed.
In total, 21 of the 26 ground-truth meshes are obtained with the Artec Spider scanner.
 For objects with narrow concavities or thin structures that are difficult to observe with optical scanning (\obj{plant}, \obj{grid}, and \obj{spine}), or for which the opacifying spray proved ineffective (\obj{paw pad} and \obj{bowl}), the ground truth is instead obtained from computed tomography scans acquired with a NAEOTOM Alpha CT scanner at a voxel size of 0.117 mm; for the \obj{warrior head}, the Artec and CT scans are fused. The scans are then segmented using Materialise Mimics 26.0 \cite{mimics}.

This workflow follows a monitored pipeline in which each reference model is inspected manually. Among the 28         objects, two do not have ground-truth scans. The first is \obj{pillow}, because its surface texture is partly interpreted by the Artec Spider as geometric variation. The second is \obj{coin}, because neither scanning modality could capture its fine geometric details reliably. Both objects are nevertheless released with their images, object masks and calibrated cameras, but they are excluded from the evaluation: all quantitative results are reported on the remaining 26 objects.

\subsection{Camera Calibration and Scan Alignment}

This section describes the reference-data pipeline used to place all images and ground-truth scans in a common metric coordinate system. We first present the camera calibration and its validation, then describe the scan-alignment procedure and its validation.

\textbf{Camera calibration. } The camera-calibration procedure is designed to be object-agnostic while providing quantified accuracy. Object-agnostic calibration is important because camera poses should not depend on the geometry, texture, or material of the reconstructed object. Otherwise, pose errors could vary with object difficulty, making reconstruction errors harder to interpret.

Several strategies are possible. One option is to rely on the robot and turntable poses, but prior work~\cite{most2025} has shown that this can be insufficiently accurate at this image resolution. Another option, adopted by EvalMVX~\cite{evalmvx}, is to place a calibration board below the object. This provides image features for pose estimation, but is not well suited to calibrated photometric stereo because the checkerboard can introduce inter-reflections and make object-contour extraction less robust during the subsequent alignment procedure. A related strategy, used in OpenSubstance~\cite{opensubstance}, places markers on a vertical cylindrical support below the object, which likely reduces inter-reflections compared with a planar checkerboard. In contrast, our calibration uses only the textured black fabric covering the turntable, with the object and background excluded from feature extraction. In our evaluation,  this provides the lowest and most consistent reprojection errors on a common set of test corners, while avoiding inter-reflections from markers.

Concretely, all images from each acquisition are calibrated jointly in a single Structure-from-Motion (SfM) run. This includes the single-light images, the multi-light images acquired with all LEDs active, and the opacifying-spray images when available. We use Segment Anything Model segmentation to generate turntable masks that restrict feature extraction to the black fabric region. SfM is performed with Meshroom~\cite{meshroom}, which builds on OpenMVG~\cite{openmvg}, using DSP-SIFT~\cite{dspsift} features. Additional details on the calibration are provided in the appendix (Section~\ref{supp:sfm}).

\textbf{Calibration accuracy.}
We compare the proposed calibration procedure with three alternatives, some of which are representative of existing benchmarks.
The first is COLMAP~\cite{colmap} applied to the same textured cloth.
The second uses a ChArUco board~\cite{charuco}, analogous to the ArUco-board calibration used with Metashape in EvalMVX~\cite{evalmvx}.
The third uses an ArUco cylinder placed on the turntable base, analogous to the ARTag fiducial markers~\cite{artag} used in OpenSubstance~\cite{opensubstance}.
We use ArUco markers~\cite{charuco} instead of ARTag markers, as they have been reported to provide better performance in prior evaluations~\cite{Patru_fiducial,sagitov_aruco}.
Accuracy is evaluated on an \(18 \times 18\)-corner checkerboard placed on the turntable, which is visible in the calibration images but excluded from feature extraction by the segmentation masks, so that its corners are never used for calibration.
We measure the reprojection error of the checkerboard corners on the 20 views with the largest viewing angle with respect to the board plane (\(55.6^\circ\)--\(89.0^\circ\)), the same selection for all four configurations, so that all of them are evaluated on the same 6\,480 corners.
Our SfM-based approach achieves the lowest reprojection error, with a mean of \(0.542\) px (median \(0.524\) px), corresponding to approximately \(0.04\) mm at the nominal object distance.
It is followed by the ArUco cylinder with a mean of \(0.666\) px (median \(0.542\) px), COLMAP with \(0.809\) px (\(0.799\) px), and ChArUco with \(1.086\) px (\(0.924\) px).
Note that the measured residual includes checkerboard-corner localisation error in addition to calibration error. Full distributions are reported in the appendix (Table~\ref{tab:per_corner_residuals}).

\textbf{Scan alignment.}
We formulate scan alignment as a contour-based ICP problem in the image domain.\footnote{Code: \url{https://github.com/martine-benchmark/contour-registration}.}
The goal is to align the ground-truth mesh with the calibrated image sequence by matching projected mesh silhouettes to object contours extracted in the images.
Since the ground-truth scans are expressed in metric units, while SfM camera poses are recovered up to an unknown scale, the alignment estimates a 7-DoF similarity transform consisting of rotation, translation, and uniform scale.
The inverse transform is then applied to the camera poses, expressing them in the metric coordinate frame of the ground-truth scan and yielding a metrically consistent image-to-geometry alignment.
Target contours are extracted from single-light images using Canny edge detection within a local search band defined from SAM object masks, with contour-extraction parameters optimised per object. For specular or transparent objects, we use images captured after applying the white sublimating opacifying spray, which increases the contrast between the object and the black turntable near the silhouette. The registration then optimises the similarity transform by minimising the Euclidean distance from the target image contour points to the projected silhouette of the ground-truth mesh. The mathematical formulation of the optimisation and the selection of the contour-extraction parameters are provided in the appendix (Section~\ref{supp:canny}).

\textbf{Alignment accuracy. } 
The alignment is validated on control 
views excluded 
from registration. For each object, we select eight control views by k-means clustering of the camera centres.
At each iteration, the similarity transform is estimated from the remaining views using the object-specific contour-extraction parameters, and the reprojection residual is measured on the excluded view.
\input{tables/loo_registration_table}We report the average residual over the eight leave-one-control-view evaluations, yielding a pixel-level image-to-geometry error at the original sensor resolution.
The mean LOO residual across all objects is 1.22 px, with a median of 1.20 px, and all 26 objects achieve a mean residual below 2 px, as shown in Table~\ref{tab:loo_registration_compact}.
We report reprojection residuals rather than silhouette IoU, as used in EvalMVX~\cite{evalmvx}, 
because preliminary tests showed that IoU on downsampled masks can be insensitive to small contour misalignments.
Additional details are provided in the appendix (Figure~\ref{fig:iou}).

\textbf{Taxonomy of reconstruction challenges.} Existing benchmarks typically report one aggregate score per object, even though a single object may contain several distinct surface properties, such as matte regions, low texture, high curvature, transparency, or specularity. As a result, an object-level score does not indicate which properties drive the reconstruction error. To make this analysis possible, we annotate each ground-truth mesh with per-vertex labels describing relevant geometric and optical properties.
The seven property labels are defined as follows.
\textit{Lambertian} denotes matte regions whose appearance is approximately diffuse.
\textit{Specular} denotes regions with strong view-dependent reflectance, including glossy or metallic surfaces.
\textit{Transparent} denotes transmissive regions, for which image evidence can be affected by refraction, reflection, or background visibility.
\textit{Textureless} denotes regions with absent surface texture, which provide very limited local image correspondences.
\textit{Fine structure} denotes thin geometric elements with small cross-section, such as wires, blades and spikes.
\textit{High curvature} denotes vertices with large local maximum principal curvature ($\vert{}k\vert{} > 0.75\text{ mm}^{-1}$), estimated using the Voronoi Covariance Measure ($R = 0.5\text{ mm}$, $r = 1\text{ mm}$) on a densified ground-truth mesh (edge splitting $> 0.15\text{ mm}$).
\textit{Low visibility} denotes vertices observed in only a small number of calibrated views, thresholded at 1 to 3 views in our experiments.
This property-level evaluation is particularly useful for regions that occupy only a small fraction of the object surface.
Errors on such under-represented regions can be diluted in global metrics, such as mean F-score or Chamfer distance over the full object.

%% file: tables/loo_registration_table.tex
\begin{wraptable}{r}{0.65\textwidth}
\vspace{-4pt}
\centering
\caption{Leave-one-out registration residuals (mean reprojection error in pixels, over 8 control views per object.
}
\label{tab:loo_registration_compact}
\scriptsize
\setlength{\tabcolsep}{1.5pt}
\begin{tabular*}{0.63\textwidth}{@{\extracolsep{\fill}} *{8}{c}}
\toprule
\obj{rock} & \obj{vase} & \obj{hammer head} & \obj{candle} & \obj{can} & \obj{glass} & \obj{petanque ball} & \obj{jam pot} \\
\midrule
1.06 & 1.08 & 1.02 & 0.78 & 1.21 & 1.14 & 0.84 & 1.26 \\
\bottomrule
\end{tabular*}
\begin{tabular*}{0.63\textwidth}{@{\extracolsep{\fill}} *{9}{c}}
\toprule
\obj{colander} & \obj{wood} & \obj{plate} & \obj{paw pad} & \obj{moka pot} & \obj{lego} & \obj{scissors} & \obj{knife} & \obj{spikes} \\
\midrule
1.11 & 1.45 & 1.24 & 1.01 & 1.98 & 1.35 & 0.89 & 1.10 & 1.11 \\
\bottomrule
\end{tabular*}
\begin{tabular*}{0.63\textwidth}{@{\extracolsep{\fill}} *{9}{c}}
\toprule
\obj{straw hat} & \obj{sneakers} & \obj{saw} & \obj{whisk} & \obj{warrior head} & \obj{plant} & \obj{grid} & \obj{bowl} & \obj{spine} \\
\midrule
1.36 & 1.27 & 1.09 & 1.18 & 1.27 & 1.31 & 1.28 & 1.69 & 1.63 \\
\bottomrule
\end{tabular*}
\\[2pt]
\vspace{-10pt}
\end{wraptable}

%% file: sec/4_experiments.tex
\section{Experiments}
\label{sec:experiments}

\subsection{Baselines}
\label{sec:baselines}

We evaluate representative methods from the reconstruction paradigms supported by the dataset.
For photogrammetry, we use Meshroom~\cite{meshroom}, an open-source pipeline based on CMPMVS~\cite{cmpmvs}, and RealityScan~\cite{realitycapture}, a widely used closed-source software package.
As a classical MVS baseline, we use COLMAP~\cite{colmap,colmap_mvs}.
We also evaluate PGSR~\cite{pgsr} for 3D Gaussian Splatting-based surface extraction, and two neural implicit surface reconstruction methods,  NeuS2~\cite{neus2} and Neuralangelo~\cite{na}.
For multi-view photometric stereo, we evaluate SuperNormal~\cite{supernormal} and RNb-NeuS2~\cite{rnbneus2}, using the state-of-the-art photometric stereo method UniMS-PS~\cite{unimsps} to estimate per-view normal maps from the multi-light images. Method settings are given in the appendix (Section~\ref{sec:supp_baselines}).
For the per-object and per-property analyses, each family is represented by one method: RealityScan, COLMAP, PGSR, NeuS2 and RNb-NeuS2. For neural implicit surface reconstruction, we decide to use NeuS2 as the representative method because Neuralangelo takes about 160 times as long to run at quarter resolution.

\subsection{Evaluation Protocol}
\label{sec:protocol}

\textbf{Image resolutions.}
Each method is evaluated at up to four image resolution levels: full resolution ($9568 \times 6376$), half resolution ($4784 \times 3188$), quarter resolution ($2392 \times 1594$), and one-eighth resolution ($1196 \times 797$).
Only the two photogrammetry pipelines (Meshroom and RealityScan) process full-resolution images: all other methods fail at this resolution for lack of GPU memory and are evaluated at lower resolutions. RealityScan is run at all four levels; COLMAP, PGSR, NeuS2 and RNb-NeuS2 are run at half, quarter and one-eighth resolution, and Neuralangelo and SuperNormal at quarter resolution only. View subsets of 40, 20, 10 and 5 views, selected by k-means, are evaluated at quarter resolution for RealityScan, COLMAP, PGSR, NeuS2 and RNb-NeuS2 (Table~\ref{tab:coverage} in the appendix).

\textbf{Evaluation protocol.}
Before evaluation, reconstructed meshes are filtered by removing vertices not visible from any camera, and unreliable regions of the ground-truth meshes (e.g., mounting bases) are excluded using per-object 3D masks; both are then cleaned and resampled identically.
Surface accuracy is measured using precision, recall, and F-score curves for distance thresholds between 0 and 4\,mm.\footnote{Evaluation code: \url{https://github.com/martine-benchmark/eval-3d}.}
Precision (accuracy) is the fraction of reconstructed points within a given distance of the reference surface, recall (completeness) the fraction of reference points within that distance of the reconstruction, and the F-score their harmonic mean.
Full details are given in Section~\ref{sec:supp_eval} of the appendix.

\subsection{Results}

Figure~\ref{fig:prf_curves} reports precision, recall, and F-score curves pooled over all objects at quarter resolution. In this configuration, the neural implicit methods reach the highest F-scores (77.9\% for Neuralangelo and 75.5\% for NeuS2 at 0.5\,mm), followed by COLMAP (70.6\%), whose precision is the highest beyond 0.6\,mm but whose recall is lower. The multi-light methods do not outperform their single-light counterparts: RNb-NeuS2 (66.7\%) has a higher recall than COLMAP but a lower precision, and remains below NeuS2, on which it builds, while SuperNormal ranks last (60.8\%). For the present benchmark and implementations, the photometric cues estimated from the multi-light images therefore do not translate into a higher global accuracy, which indicates that further work is needed to fully exploit photometric-stereo information in multi-view surface reconstruction.

\begin{figure*}[t]
  \centering
  \includegraphics[width=\textwidth]{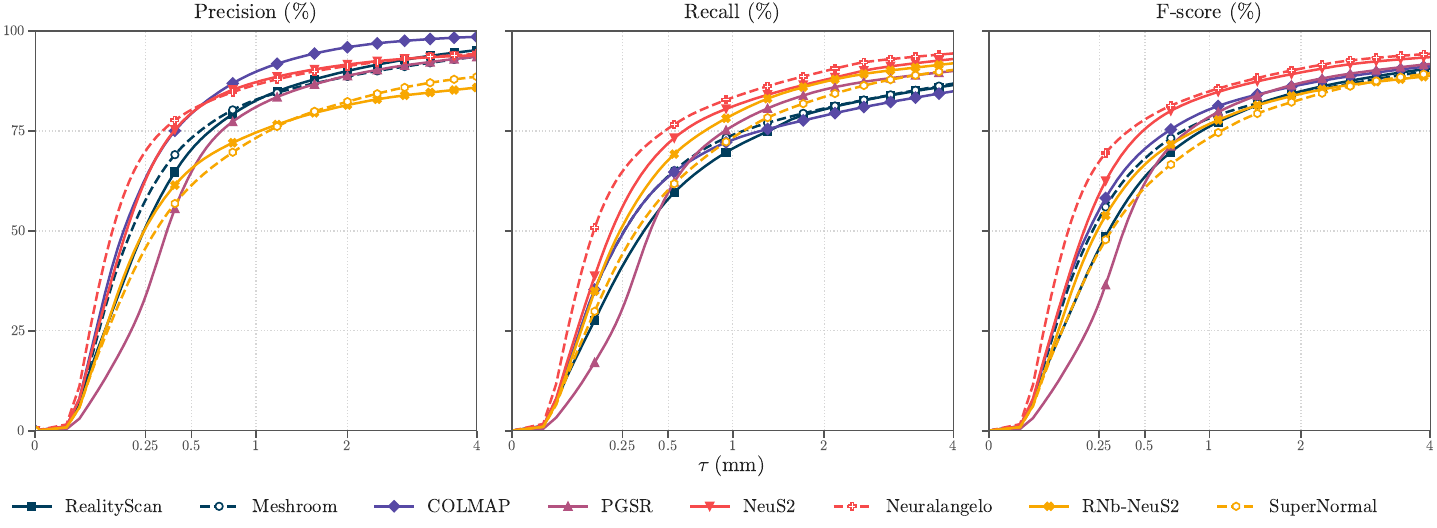}
  \caption{Global precision, recall, and F-score curves pooled across all evaluated objects at quarter resolution (d4). Each curve aggregates all point-to-surface distances before computing the metric, so larger objects contribute proportionally more.
  }
  \vspace{-10pt}
  \label{fig:prf_curves}
\end{figure*}

\begin{wrapfigure}{r}{0.35\textwidth}
  \vspace{1pt}
  \centering
  \includegraphics[width=0.34\textwidth]{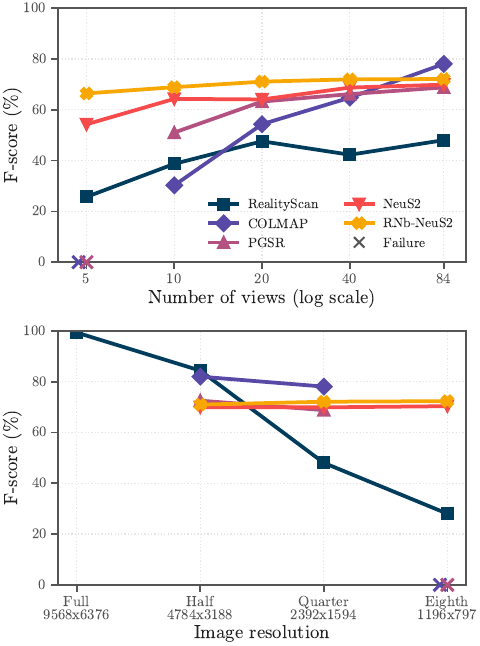}
  \caption{F-score at 1 mm on the \obj{vase} object versus number of input views on a log scale (top) and image resolution (bottom).
  Crosses plotted at 0\% F-score mark failed reconstructions in the 5-view and one-eighth-resolution settings.
  }
  \label{fig:resolution_views}
  \vspace{0pt}
\end{wrapfigure}

Figure~\ref{fig:resolution_views} studies the effect of reducing the number of views and the input resolution on the \obj{vase} object; it illustrates the per-object analyses enabled by the benchmark rather than supporting general conclusions. On this object, COLMAP and PGSR degrade under viewpoint sparsity and fail with 5 views, whereas NeuS2 and RNb-NeuS2 remain more stable, RNb-NeuS2 performing best in the sparse-view setting, possibly thanks to the normal cues provided by photometric stereo. Reducing the image resolution mostly affects the photogrammetry and MVS pipelines: RealityScan drops from 99.3\% at full resolution to 28.0\% at one-eighth resolution, and COLMAP fails at one-eighth resolution.
PGSR also fails at one-eighth resolution, whereas NeuS2 and RNb-NeuS2 are almost unaffected. Across the 26 objects (Table~\ref{tab:coverage} in the appendix), COLMAP and PGSR reconstruct 14 and 13 objects with 10 views and 4 and 3 with 5 views, while NeuS2 and RNb-NeuS2 reconstruct all of them in every configuration.

Additional exemplar results are shown in Figure~\ref{fig:qualitative}, which provides spatial error maps for representative objects and methods.

The property-level radar plot in Figure~\ref{fig:radar_chart} compares the representative methods on the same ground-truth surface samples for each property. Transparent and low-visibility regions yield the lowest F-scores for all methods except Neuralangelo, which scores lowest on low-visibility regions and fine structures (38.2\%). NeuS2 leads on specular and transparent regions, RNb-NeuS2 on high-curvature and low-visibility regions, and Neuralangelo on Lambertian and textureless ones, narrowly ahead of COLMAP on the latter (65.3\% vs.\ 63.5\%). NeuS2 and RNb-NeuS2 are on par on fine structures (92.1\% and 91.0\%), well above COLMAP (59.6\%) and RealityScan (52.3\%).
These observations should be read with caution: several gaps are small, and the analysis is conducted at a single resolution. Patch-based MVS such as COLMAP relies on local photo-consistency across views, so it is expected to be strongest where this evidence is stable and to struggle where the assumption is challenged, as on specular or transparent surfaces. Neural surface methods instead optimise a global, regularised volumetric representation, which better tolerates the view-dependent appearance of such regions, and their pixel-wise optimisation has been shown to better preserve thin structures and high-frequency detail~\cite{neus,yariv2021volume}. Confirming these trends requires testing at different resolutions and view counts. The proposed benchmark is designed for such comparisons.

\input{figures/qualitative_results}

%% file: figures/qualitative_results.tex
\begin{figure*}[t]
\centering
\setlength{\tabcolsep}{1pt}
\renewcommand{\arraystretch}{0.5}
\newcommand*{\rowlabel}[3]{\raisebox{\dimexpr(0.14\textwidth*#1/#2-\height)/2\relax}{\rotatebox{90}{\small\obj{#3}}}}
\begin{tabular}{c c c c c c c}
 & Input & RealityScan & COLMAP & PGSR & NeuS2 & RNb-NeuS2 \\
\rowlabel{1020}{576}{hammer head} &
  \includegraphics[width=0.14\textwidth]{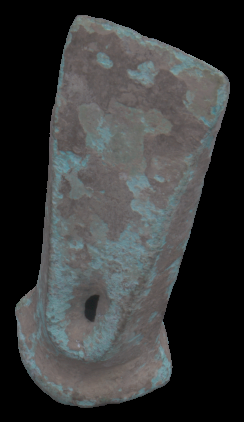} &
  \includegraphics[width=0.14\textwidth]{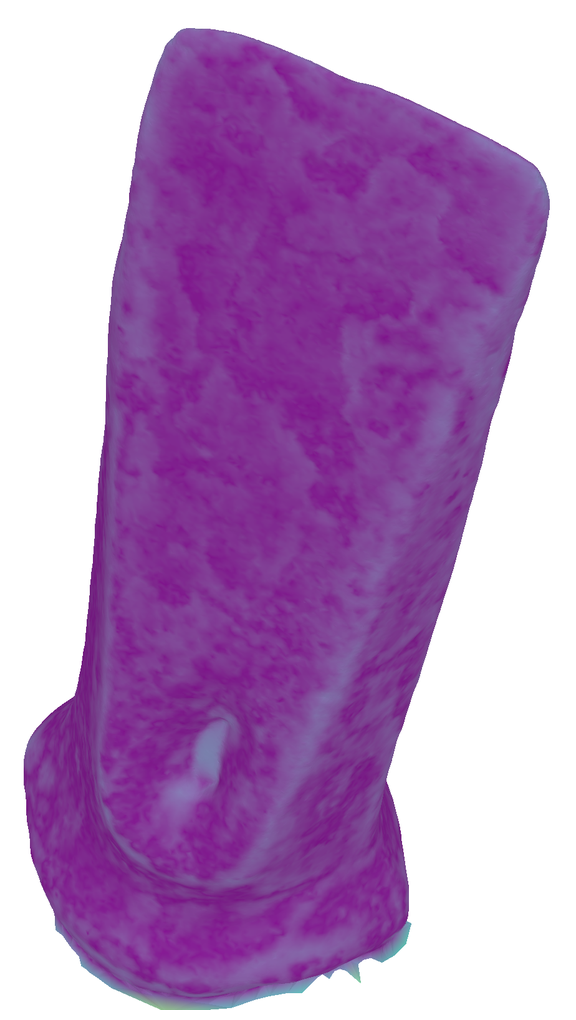} &
  \includegraphics[width=0.14\textwidth]{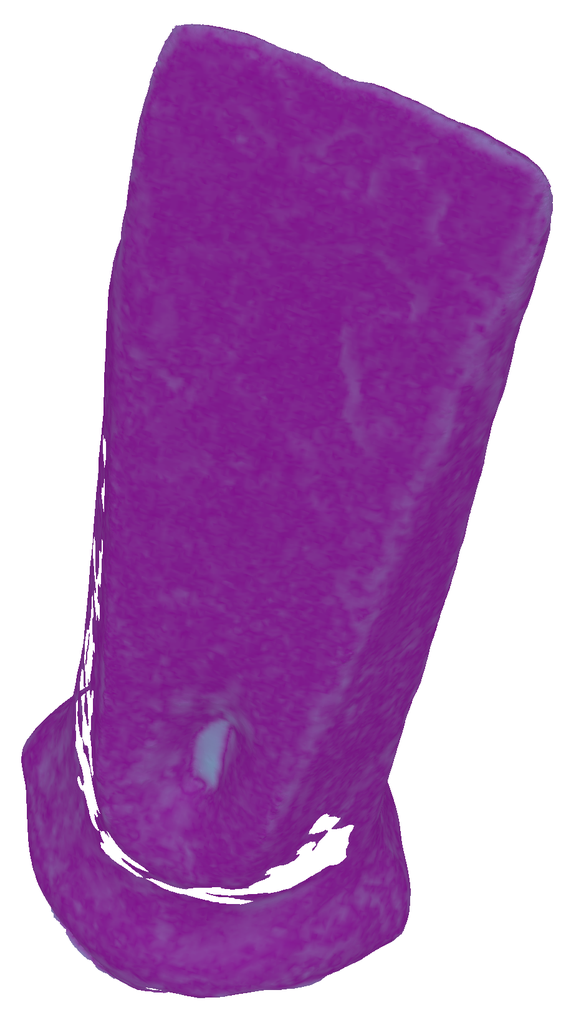} &
  \includegraphics[width=0.14\textwidth]{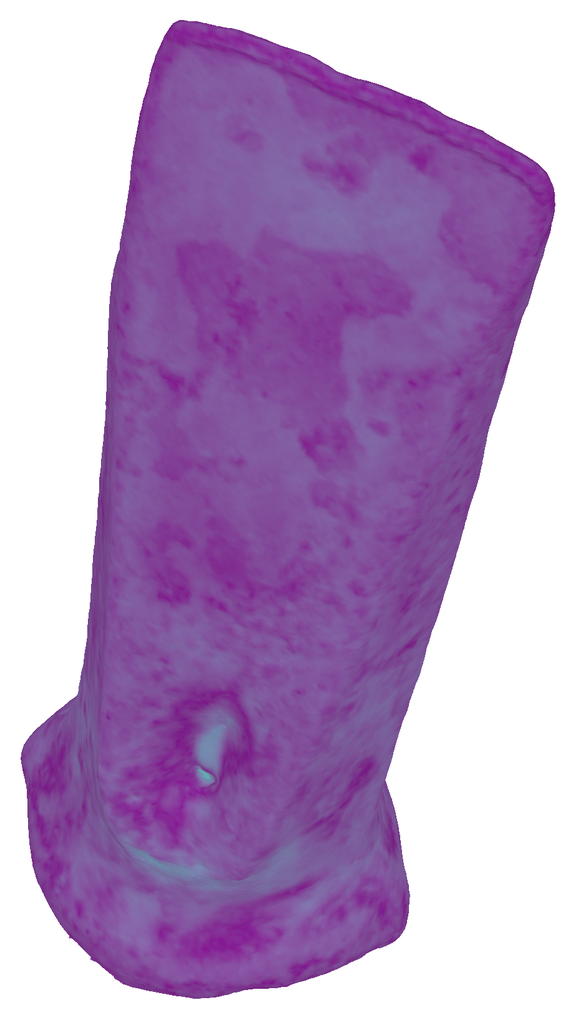} &
  \includegraphics[width=0.14\textwidth]{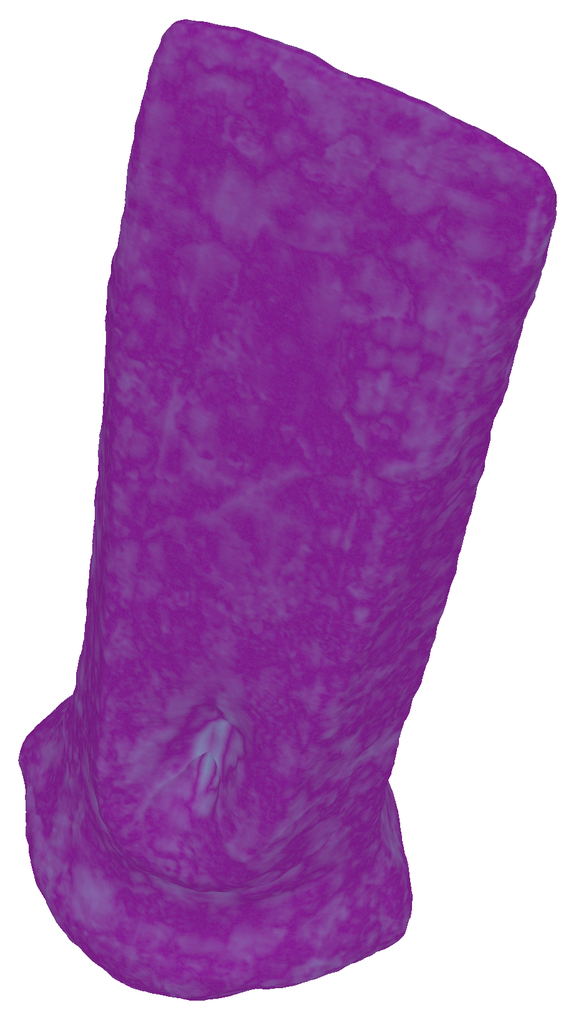} &
  \includegraphics[width=0.14\textwidth]{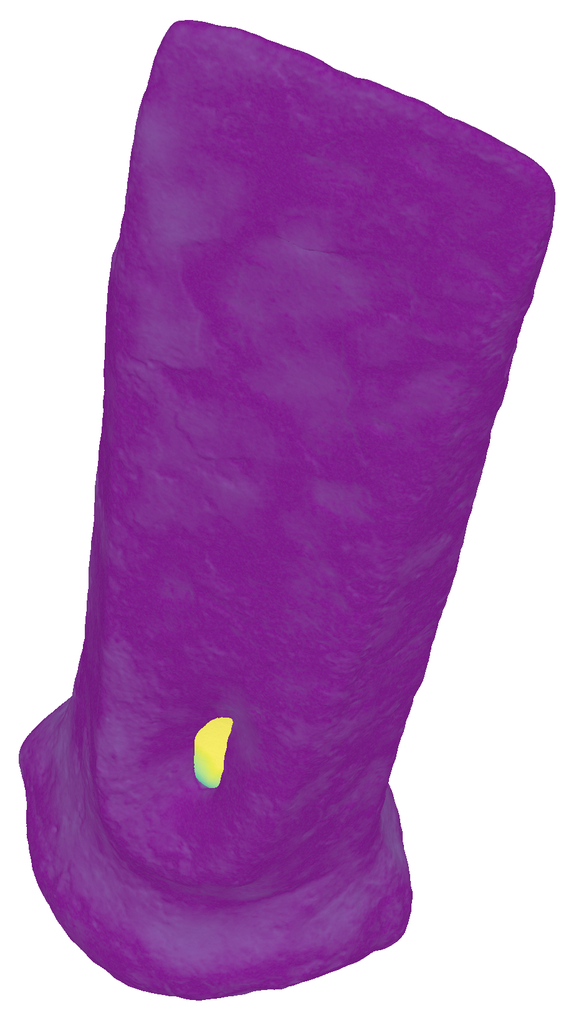} \\
\rowlabel{726}{1020}{saw} &
  \includegraphics[width=0.14\textwidth]{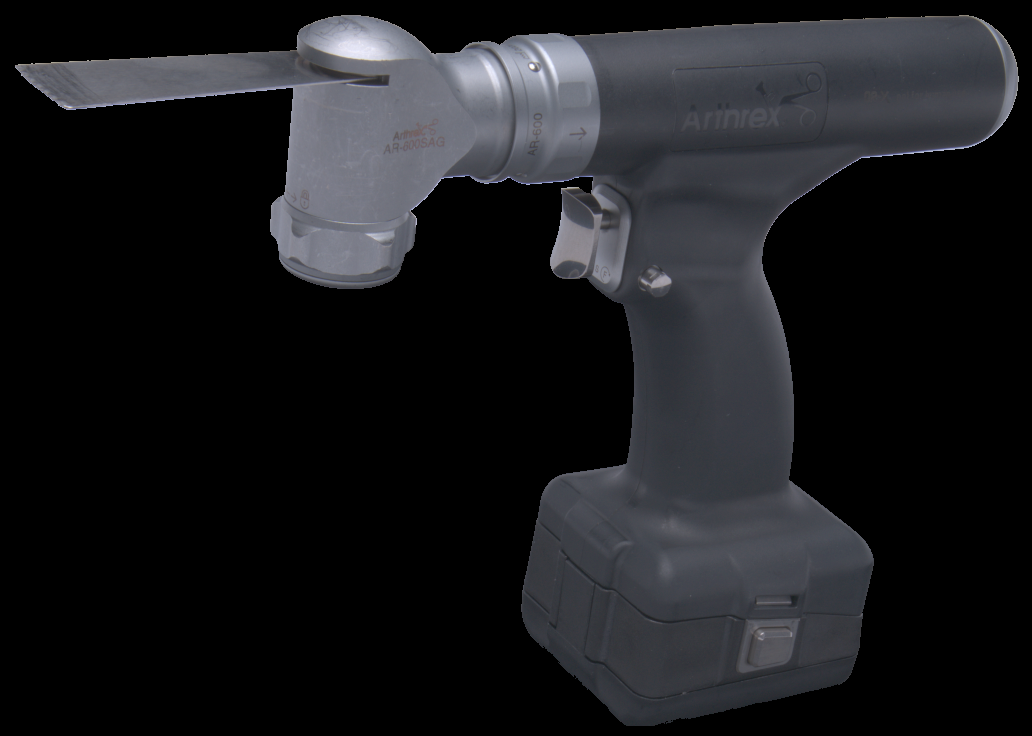} &
  \includegraphics[width=0.14\textwidth]{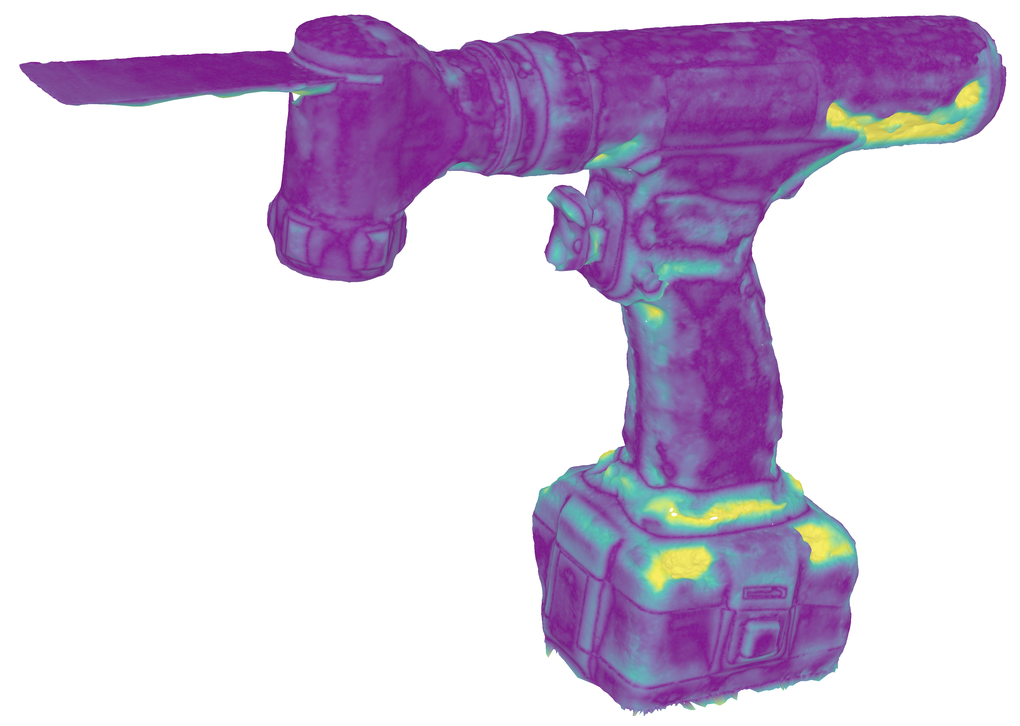} &
  \includegraphics[width=0.14\textwidth]{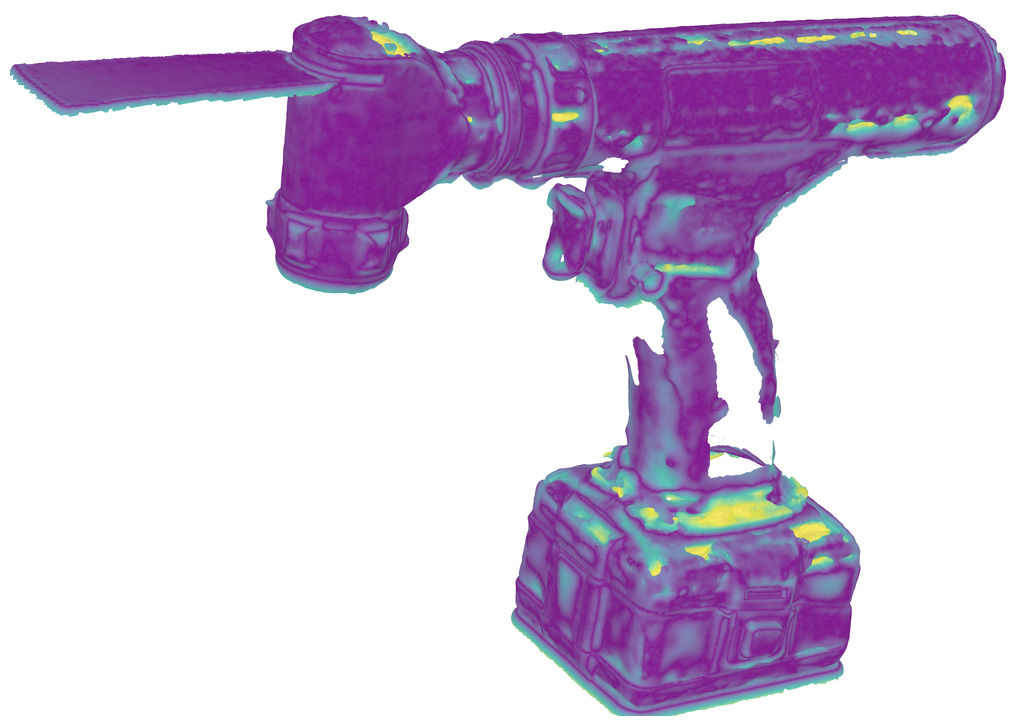} &
  \includegraphics[width=0.14\textwidth]{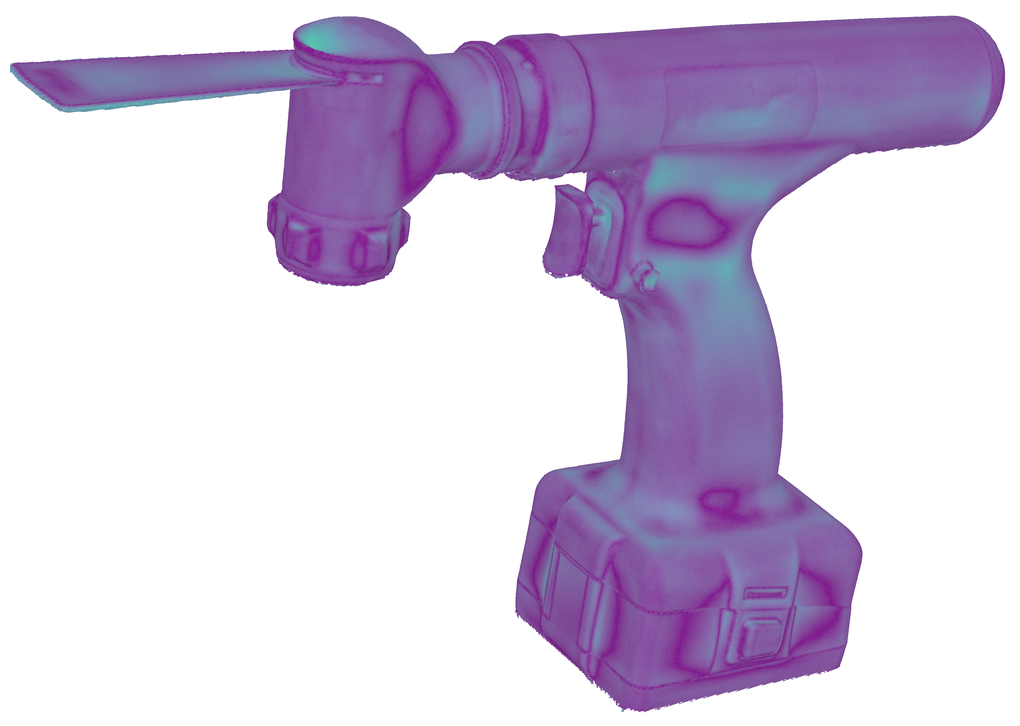} &
  \includegraphics[width=0.14\textwidth]{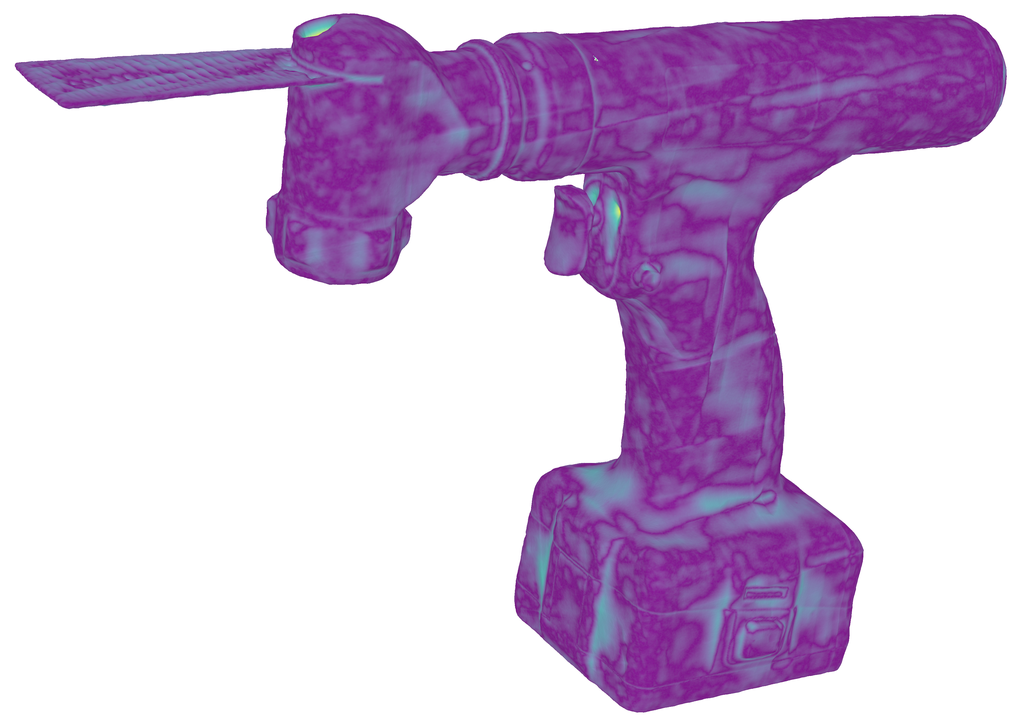} &
  \includegraphics[width=0.14\textwidth]{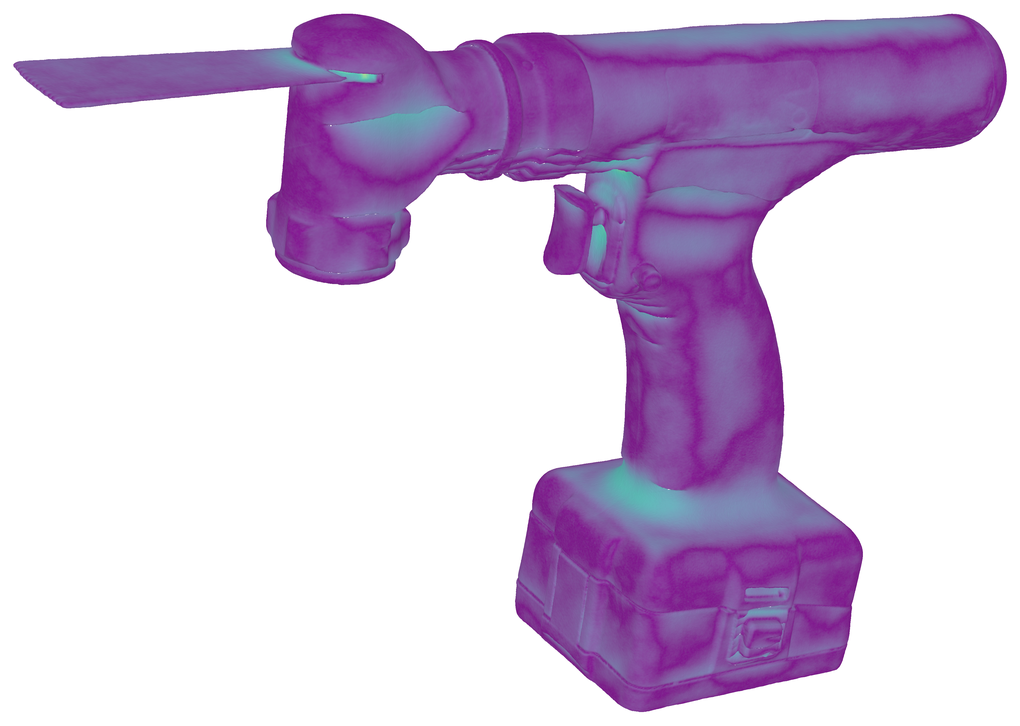} \\
\rowlabel{1020}{635}{glass} &
  \includegraphics[width=0.14\textwidth]{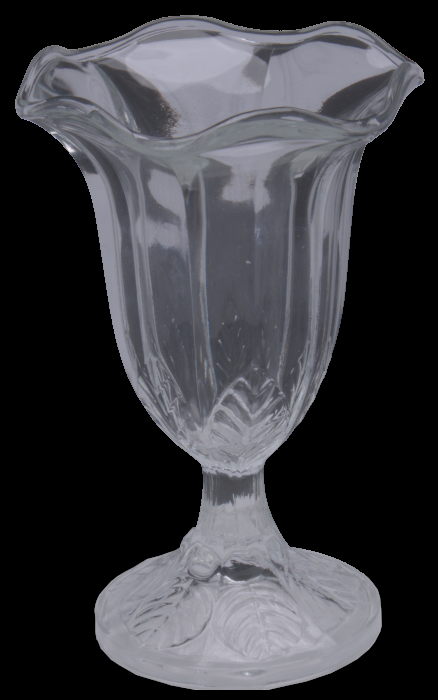} &
  \includegraphics[width=0.14\textwidth]{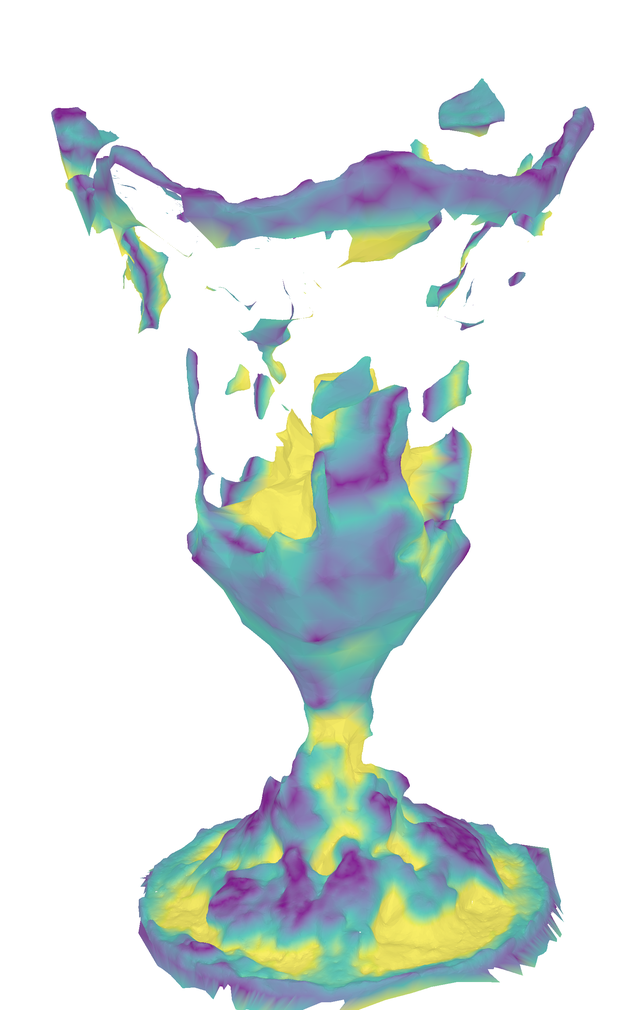} &
  \includegraphics[width=0.14\textwidth]{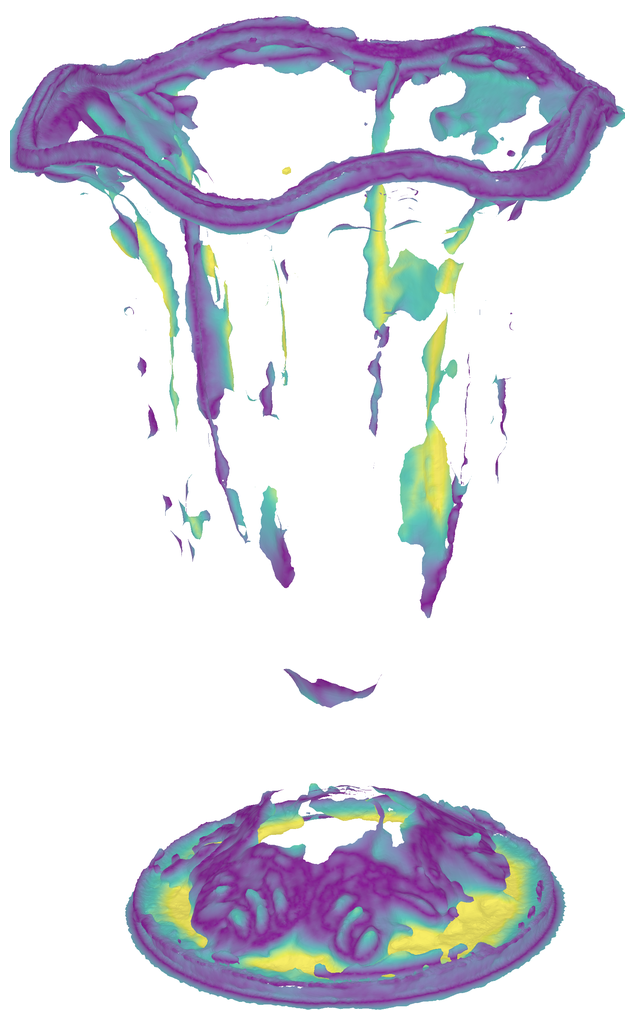} &
  \includegraphics[width=0.14\textwidth]{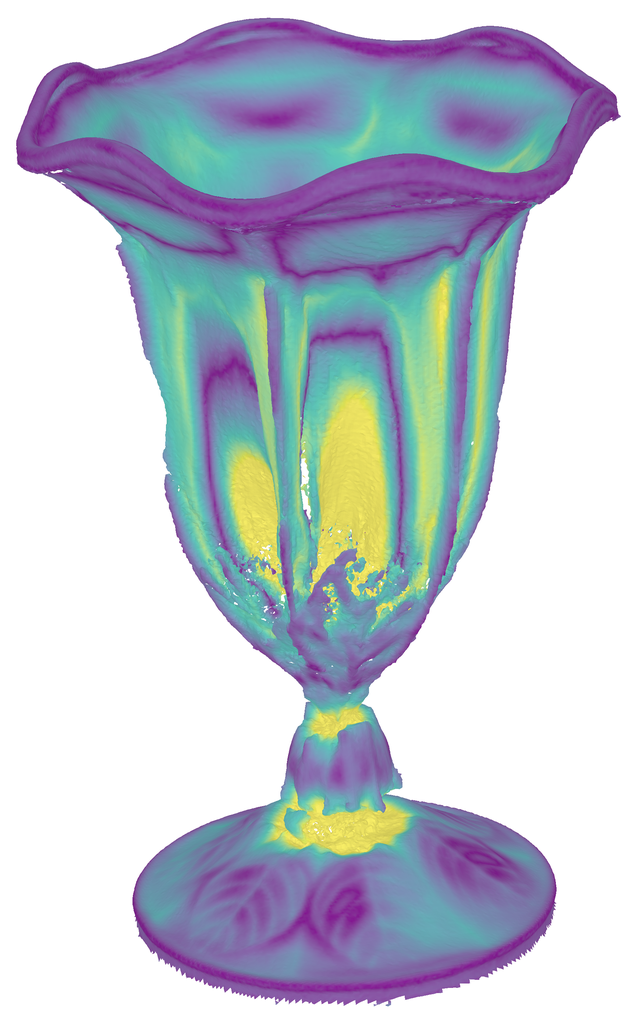} &
  \includegraphics[width=0.14\textwidth]{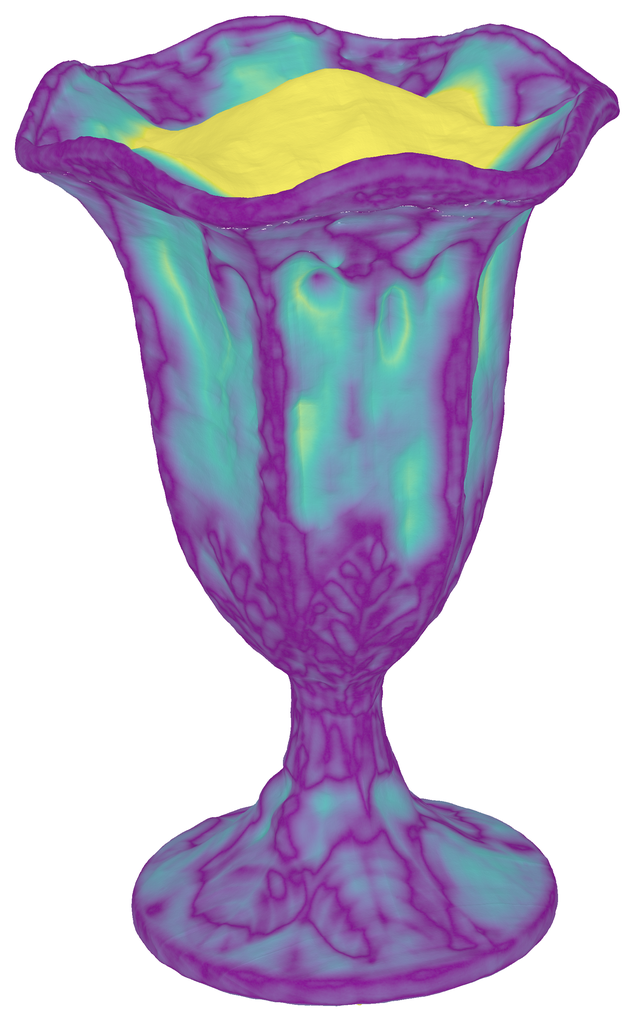} &
  \includegraphics[width=0.14\textwidth]{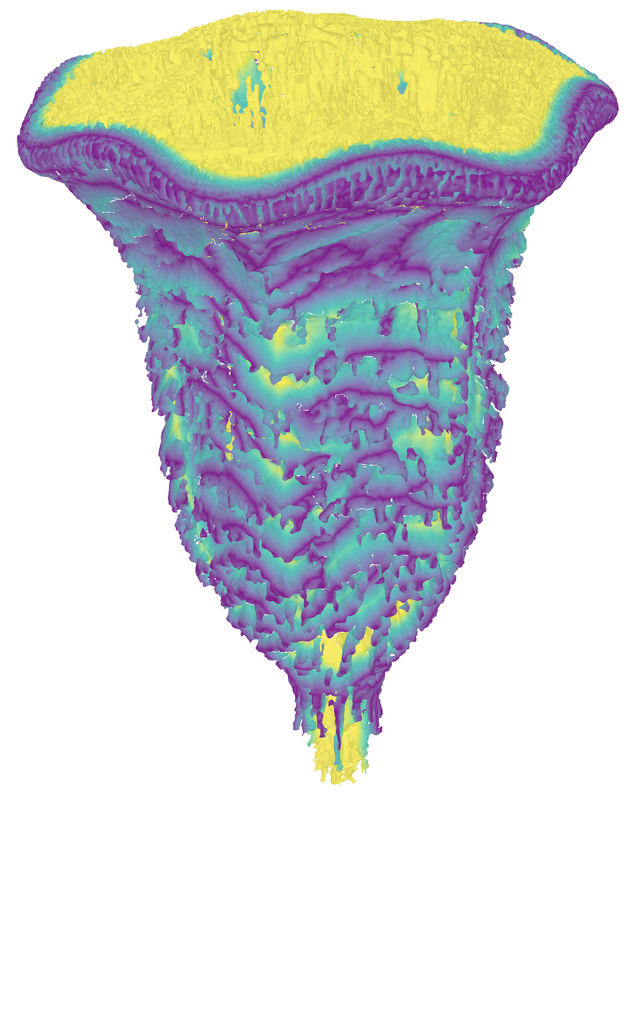} \\
\rowlabel{857}{1019}{lego} &
  \includegraphics[width=0.14\textwidth]{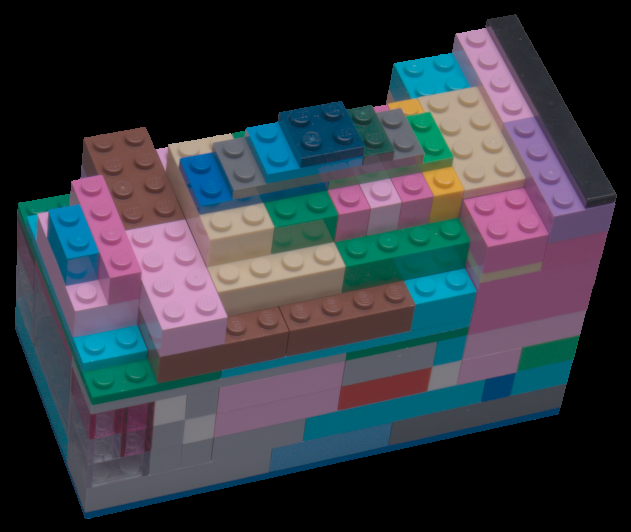} &
  \includegraphics[width=0.14\textwidth]{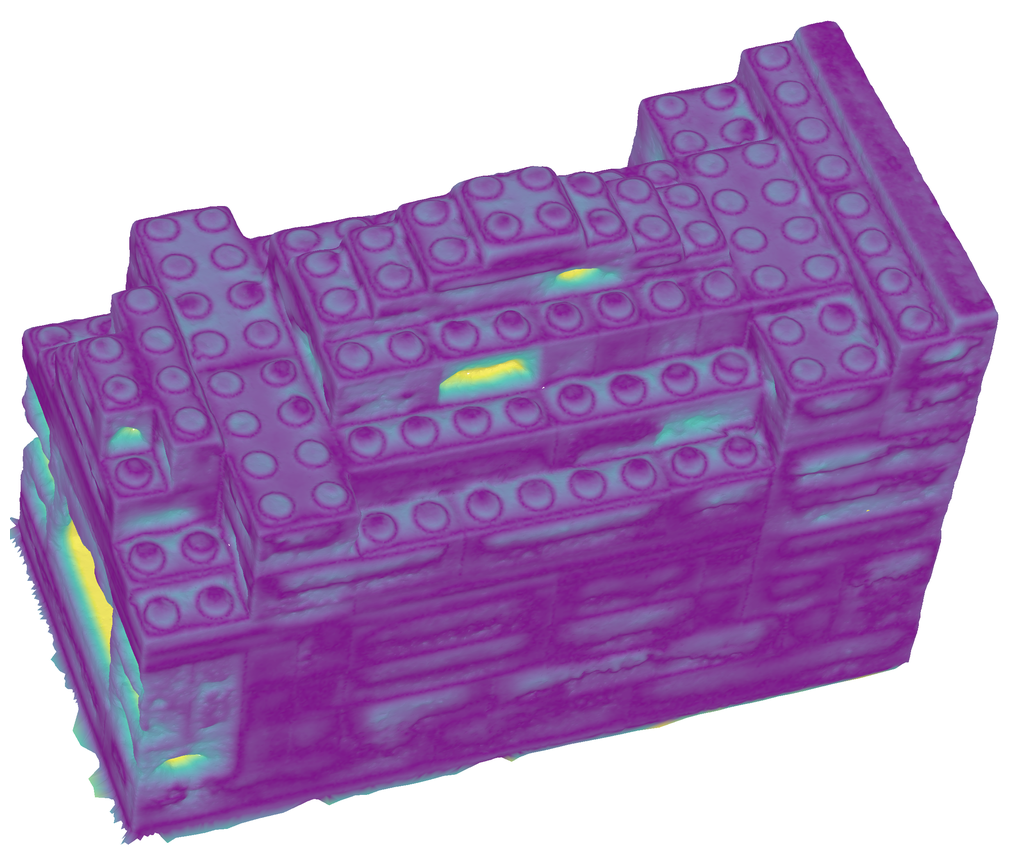} &
  \includegraphics[width=0.14\textwidth]{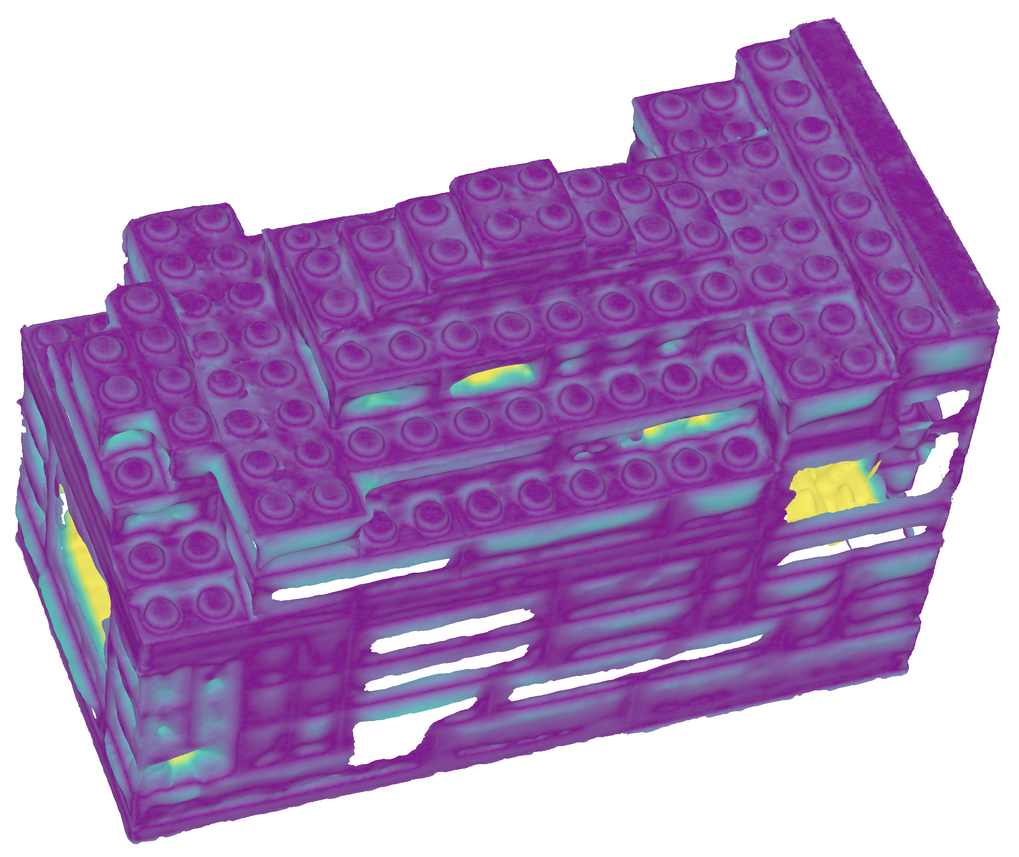} &
  \includegraphics[width=0.14\textwidth]{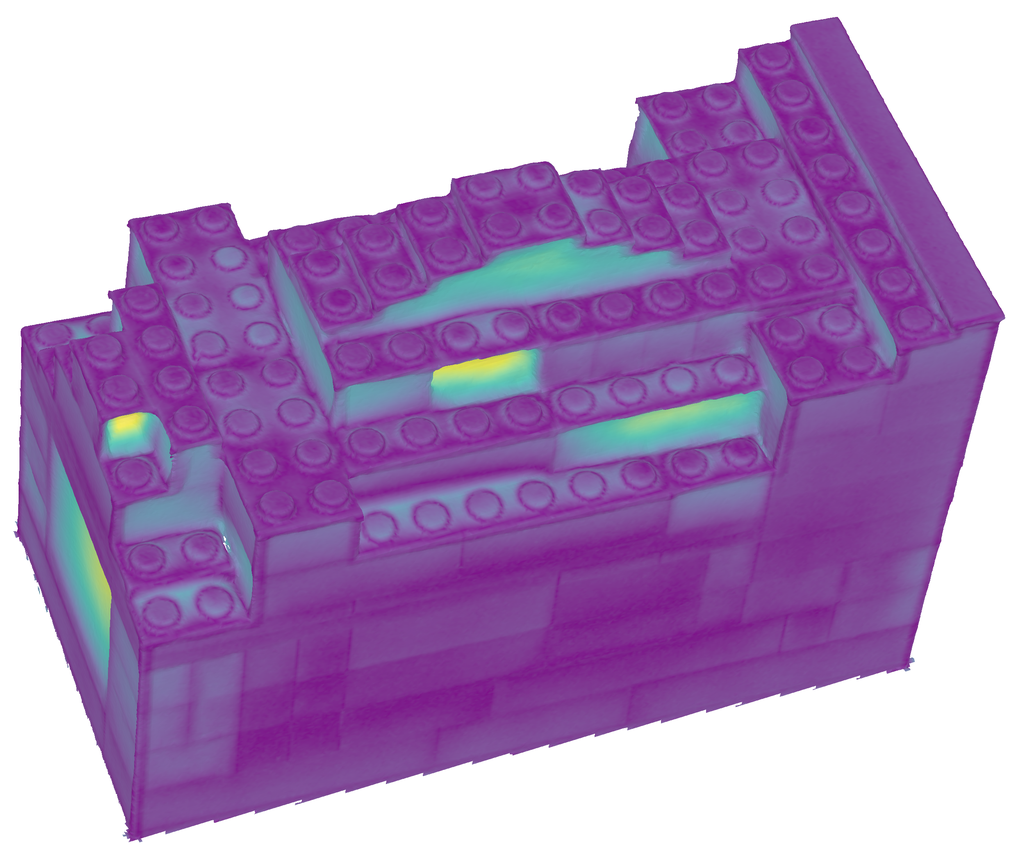} &
  \includegraphics[width=0.14\textwidth]{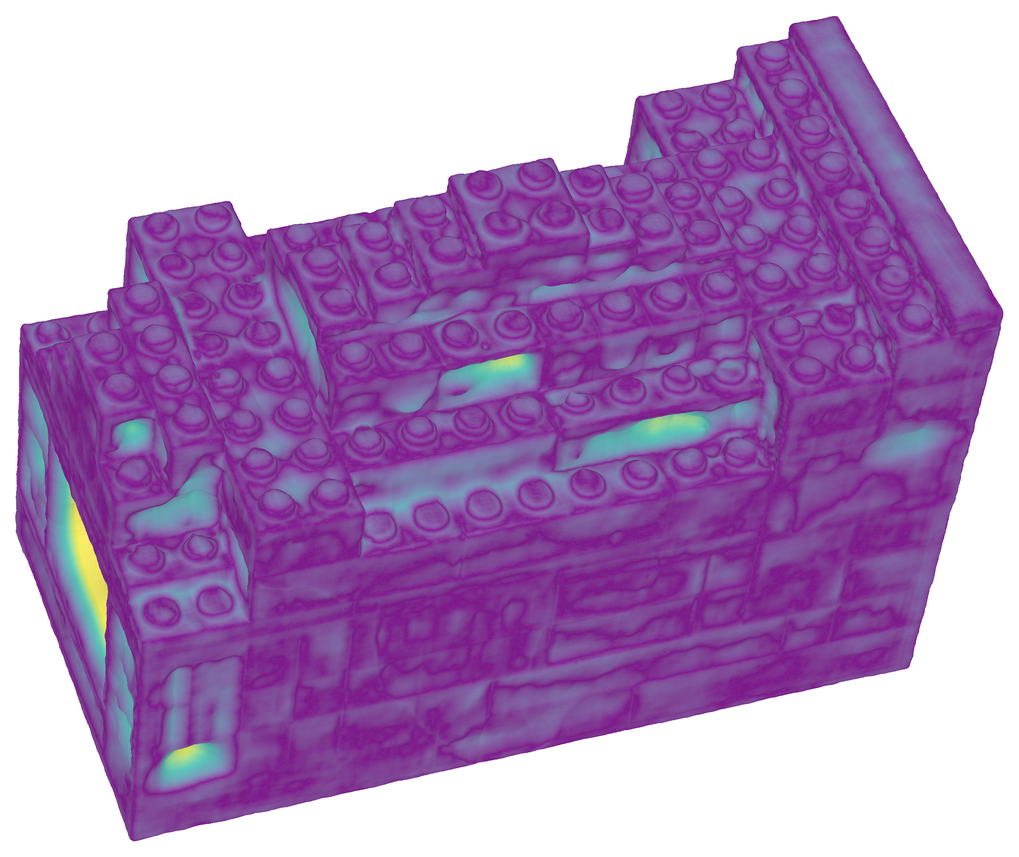} &
  \includegraphics[width=0.14\textwidth]{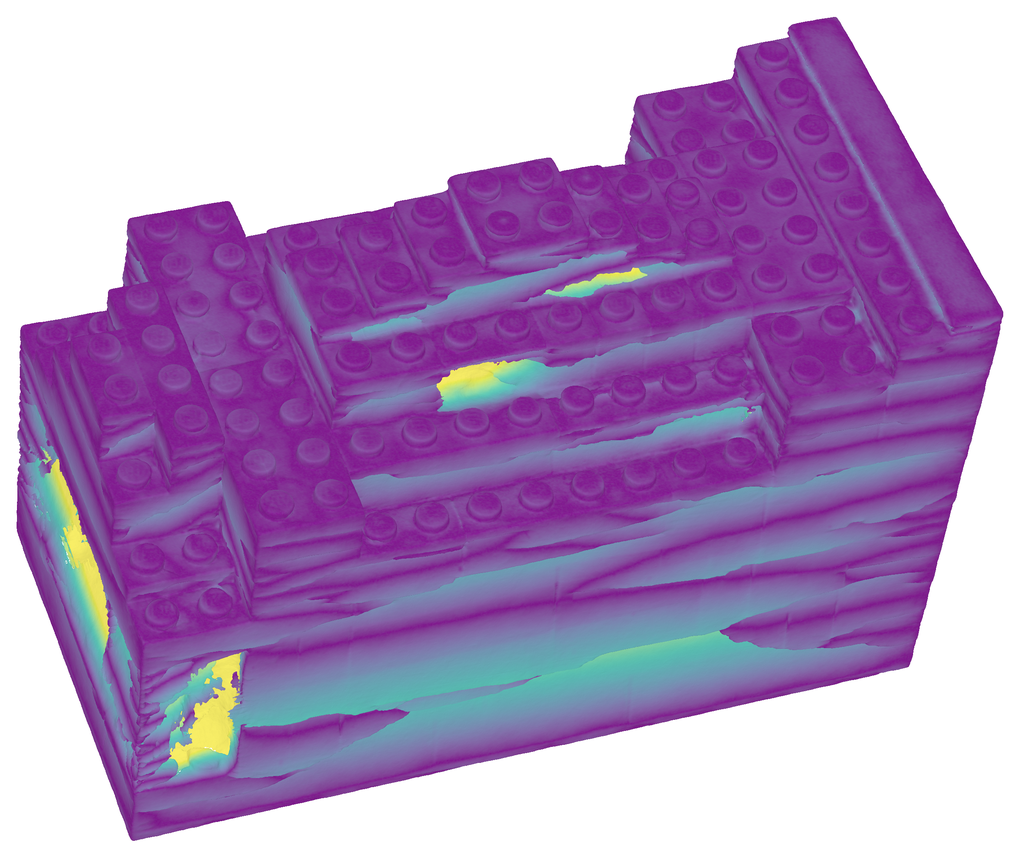} \\
\end{tabular}
\\[2pt]
\includegraphics[width=0.72\textwidth]{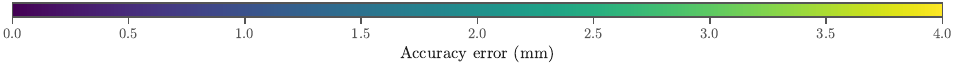}
\caption{Qualitative comparison of accuracy error heatmaps (viridis colour map, 0--4\,mm, larger errors saturated) at quarter resolution (d4). 
}
\label{fig:qualitative}
\vspace{-18pt}%
\end{figure*}%

%% file: sec/5_conclusion.tex
\section{Conclusion and Future Work}
\label{sec:conclusion}

In this work, we introduced \martine{}, a benchmark for object surface reconstruction in controlled environments designed to address limitations of existing datasets. \martine{} provides 28 objects, including challenging cases such as transparent materials and fine geometric structures, and controls acquisition parameters such as viewpoint distribution, image resolution, and illumination. 
This enables reconstruction paradigms to be compared on the same objects under well-defined acquisition regimes. 
The emphasis on reference-data accuracy, 
through object-independent calibration and quantified image-to-geometry residuals, supports a clearer interpretation of reconstruction errors.
In addition, 3D property masks enable a more detailed analysis of errors across surface regions and object properties.

\begin{wrapfigure}{r}{0.38\textwidth}
  \centering
  \includegraphics[width=0.43\textwidth]{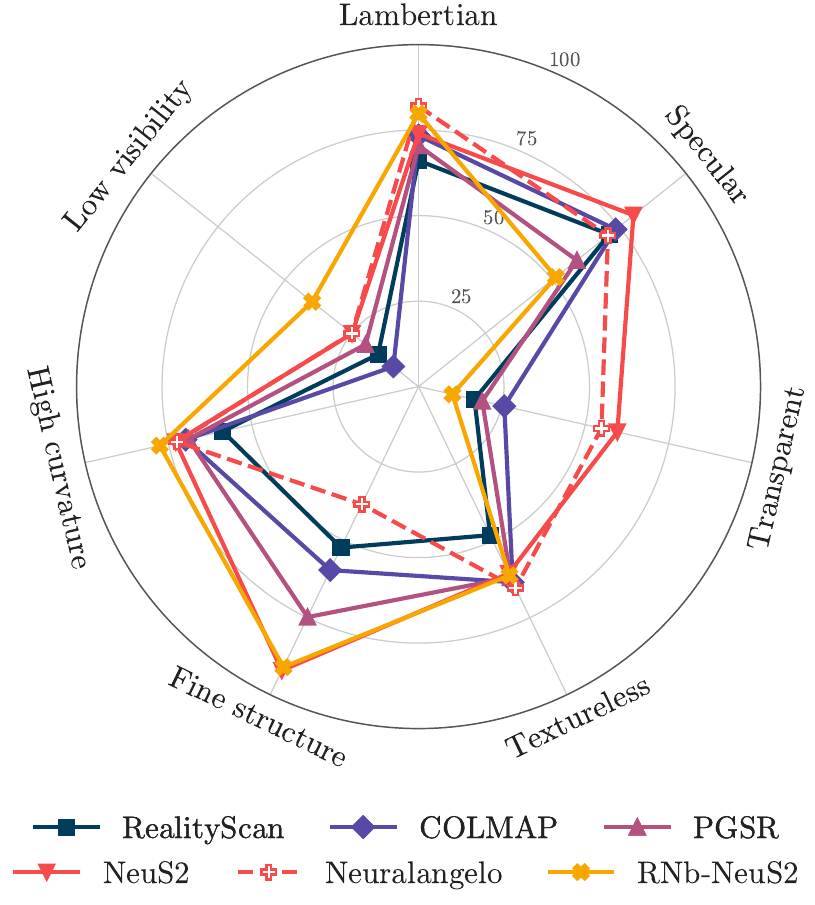}
  \caption{Radar chart of F-score at 0.5\,mm per reconstruction challenge, at quarter resolution, for the five representative methods and Neuralangelo. Higher is better.
  }
  \label{fig:radar_chart}
  \vspace{-15pt}
\end{wrapfigure}

The current version of the dataset nevertheless has limitations. First, our property taxonomy is partly qualitative, and the manual masks label only unambiguous regions. Future work could use objective criteria to automate annotation and cover a wider range of surfaces. Where a property can be measured on a continuous scale, evaluation could show how reconstruction performance varies across that scale. Second, this work focuses on surface reconstruction. However, the dataset is directly usable for novel-view synthesis, relighting, and inverse rendering, even though we do not evaluate these tasks. Third, the time required to acquire and validate the data limits the benchmark to 28 objects. Each object is captured in 2\,604 images at 61 megapixels, and the full workflow takes about one day per object, including image acquisition, scanning, alignment, and quality control. We adopted this protocol to prioritise the accuracy of the reference data.

More generally, we hope that this benchmark will encourage the community to characterise reconstruction methods by the technical challenges they address, rather than only through aggregate rankings. We also hope it will promote more systematic reporting of ground-truth accuracy, camera-calibration accuracy, and image-to-geometry alignment errors for this type of benchmark.

%% file: sec/6_supplementary.tex
\appendix

\section{Camera Calibration Specifications}
\label{supp:sfm}

\subsection{Calibration Accuracy}

\begin{figure}[ht]
    \centering
    \begin{subfigure}[b]{0.32\textwidth}
        \centering
        \includegraphics[width=\textwidth]{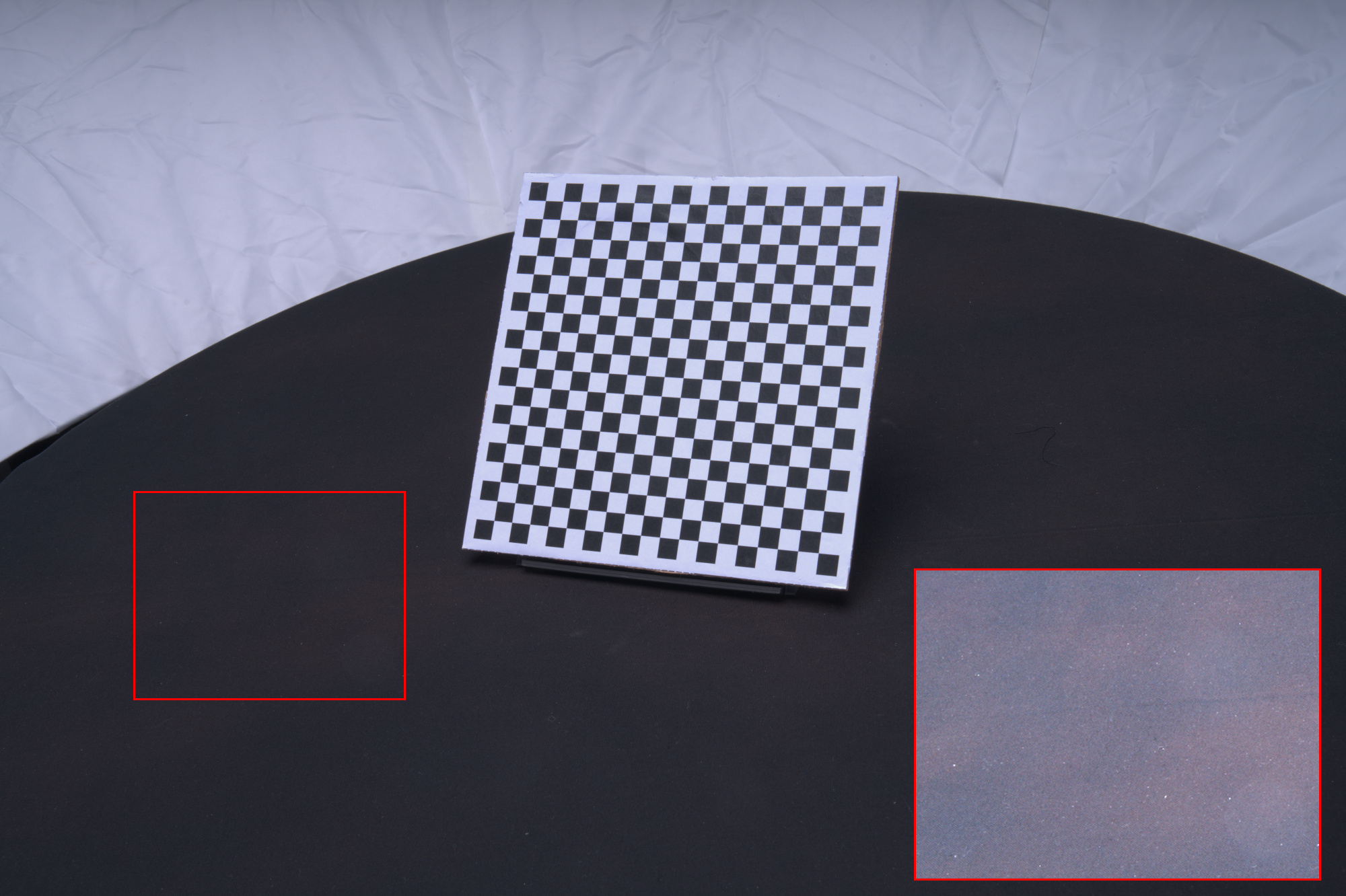}
        \caption{Textured cloth (ours)}
        \label{fig:cal_baseline}
    \end{subfigure}
    \hfill
    \begin{subfigure}[b]{0.32\textwidth}
        \centering
        \includegraphics[width=\textwidth]{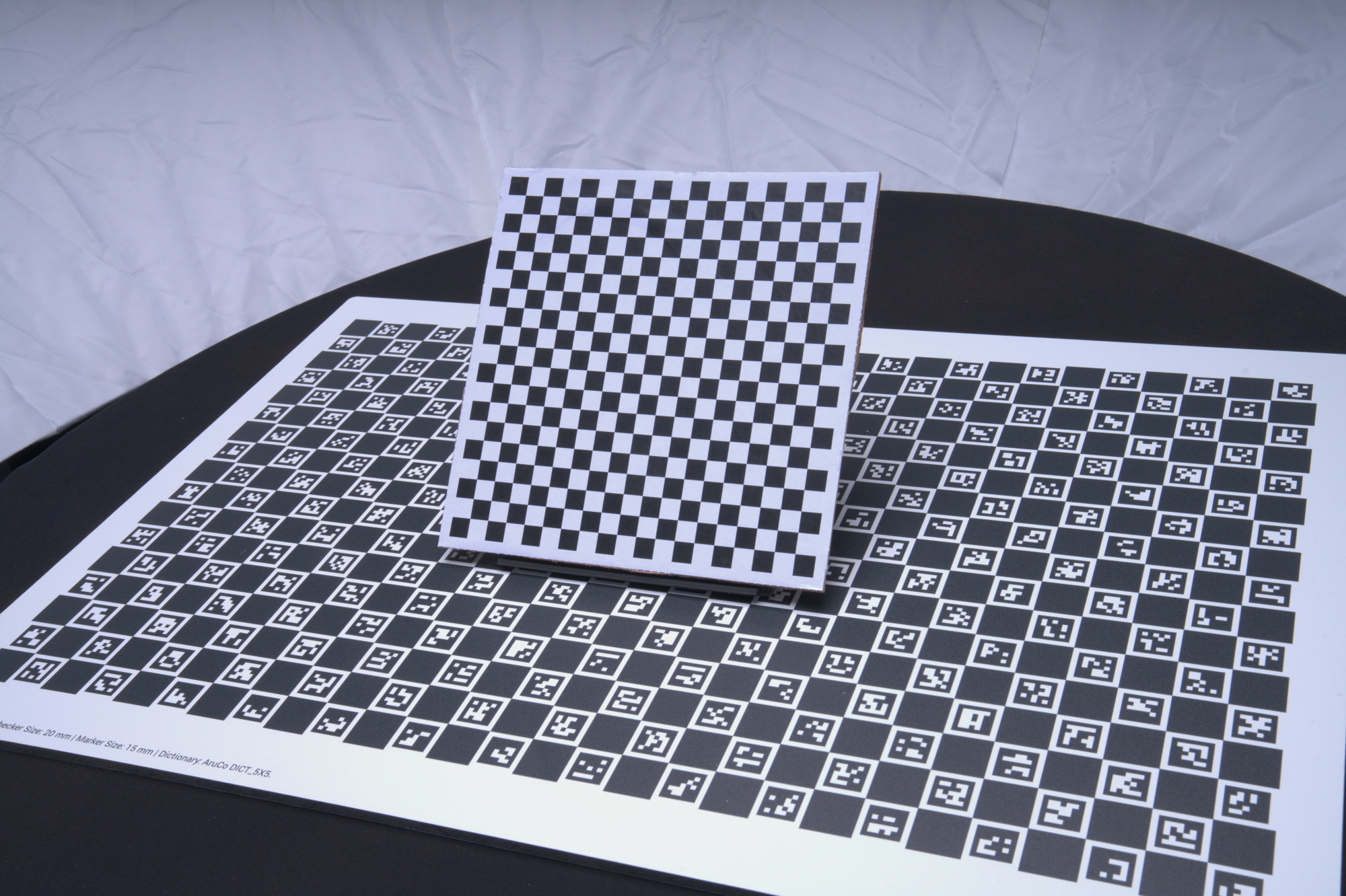}
        \caption{ChArUco board}
        \label{fig:cal_charuco}
    \end{subfigure}
    \hfill
    \begin{subfigure}[b]{0.32\textwidth}
        \centering
        \includegraphics[width=\textwidth]{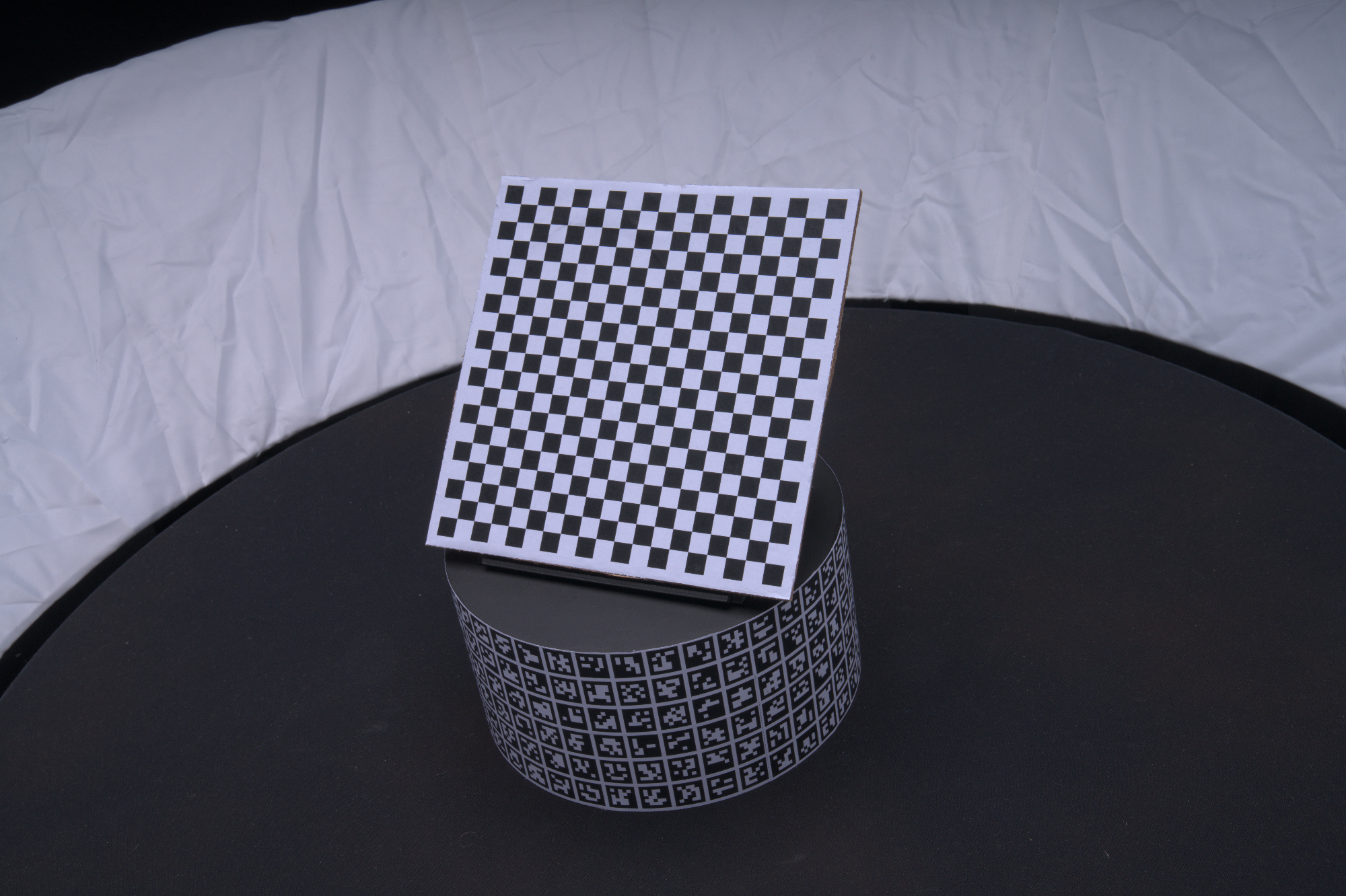}
        \caption{ArUco cylinder}
        \label{fig:cal_aruco}
    \end{subfigure}
    \caption{Calibration setup comparison. All configurations include a test checkerboard used only for evaluation. (a)~Our approach uses SfM on the textured black cloth. The inset shows a brightness-enhanced crop of the cloth micro-texture, which provides point correspondences at the ultra-high image resolution. (b)~ChArUco board \cite{calibio}. (c)~ArUco cylinder on the turntable base.}
    \label{fig:cal_setups}
\end{figure}

We compare our SfM-based camera calibration approach against three alternatives representative of existing benchmarks: COLMAP, a ChArUco board (similar to EvalMVX \cite{evalmvx}), and an ArUco cylinder (similar to OpenSubstance \cite{opensubstance}). As discussed in the main paper, accuracy is measured by the reprojection error on a test checkerboard that appears in the calibration images but is excluded from feature extraction by the segmentation masks. Figure~\ref{fig:cal_setups} illustrates these different calibration scenarios.

Table~\ref{tab:per_corner_residuals} reports the distribution of the per-corner residuals using the 20 views with the largest viewing angle with respect to the board plane (\(55.6^\circ\)--\(89.0^\circ\)). Across all four configurations, the residuals are computed over the same 6\,480 corners.
Our method obtains the lowest mean (0.542 px, about 0.04 mm) and the shortest tail (95th percentile 0.931 px, maximum 1.489 px), whereas its median is close to that of the ArUco cylinder (0.524 against 0.542 px).

\input{tables/per_corner_residuals}

\subsection{SfM Configuration Details}

Camera calibration is performed with Meshroom~\cite{meshroom} using DSP-SIFT~\cite{dspsift} features, which improve robustness to scale changes compared to standard SIFT by pooling descriptors across multiple domain sizes.
Meshroom is used with the \texttt{ultra} preset and the SfM matching is configured in exhaustive mode with the following parameters: distance ratio~$= 0.75$, maximum reprojection error~$= 1.0$~px. Meshroom is run directly on the native 16-bit EXR images, preserving the full radiometric precision of the acquisition and avoiding the degradation of sub-pixel feature localisation that can result from 8-bit quantisation. The resulting bundle adjustment achieves a root-mean-square reprojection error of 0.095 px, used here as an internal consistency metric.

\section{Segmentation \& Alignment Specifications}
\label{supp:canny}

\subsection{Canny Explanation}

To establish accurate 2D--3D registration, we employ a contour-based alignment protocol that refines initial segmentation boundaries through optimised edge detection.

\textbf{Contour extraction.} 
For each view $i$, target contour points $\mathcal{C}_i$ are extracted from the single-light images captured under rotation-invariant ambient illumination. For highly specular objects, images acquired with sublimating opacifying spray are used instead to ensure clean silhouettes.
For the \obj{knife}, we obtain the best alignment by jointly registering the scan to the single-light and multi-light images.
A search band of $m=20$ pixels is constructed around the initial Segment Anything Model (SAM) mask boundary by morphological dilation and erosion:
\begin{equation}
    \mathcal{M}_i = \mathrm{dilate}(S_i, m) \setminus \mathrm{erode}(S_i, m),
\end{equation}
and the target contour is defined as $\mathcal{C}_i = \mathrm{Canny}(I_i) \cap \mathcal{M}_i$, retaining only edge pixels that fall within this band.

\textbf{Registration objective.} 
The alignment minimises the sum of nearest-neighbour distances from the target Canny contour points to the projected silhouette vertices of the ground-truth mesh:
\begin{equation}
\label{eq:contour_icp}
\min_{R,\,\mathbf{t},\,s} \;\sum_{i=1}^{N} \sum_{\mathbf{c} \in \mathcal{C}_i} \min_{j \in \mathcal{S}(M,\,C_i)} \left\| \mathbf{c} - \pi\!\left(R\,(s\,\mathbf{x}_j) + \mathbf{t},\, K_i\right) \right\|,
\end{equation}
where $\mathbf{x}_j$ are vertices of the ground-truth mesh and $\mathcal{S}(M, C_i)$ is the set of silhouette vertices visible from camera $C_i$. The set $\mathcal{S}$ includes edges whose adjacent faces have opposite front/back visibility, filtered by rasterisation to retain only occluding contour points. The perspective projection $\pi(\cdot, K_i)$ accounts for radial distortion, and the nearest projected silhouette vertex of each contour point is found with a KD-tree. The solver uses the Trust Region Reflective method with a soft-$\ell_1$ loss for robustness to spurious Canny edges, in a coarse-to-fine scheme with outlier trimming.\footnote{Implementation details are available in the released registration code.}

\textbf{Threshold optimisation.} 
The Canny edge detection thresholds are selected per object from a grid of 20 configurations: $\tau_\ell \in \{5, 10, 20, 50\}$ and $\tau_h \in \{20, 30, 40, 60, 80, 120\}$ with the constraint $\tau_\ell < \tau_h$, a Sobel kernel size of 15, and a search margin of 20~px. For each configuration, a registration is performed and the result is evaluated by rasterising the ground-truth mesh at the registered pose to produce a reference silhouette $S$ for each view. 

For a given tolerance $\varepsilon$ (in pixels), the evaluation metrics are aggregated across views:
\begin{itemize}
    \item \textbf{Silhouette coverage} $T_R = |\{s \in S : d(s, C) \leq \varepsilon\}| \,/\, |S|$: fraction of the GT silhouette covered by Canny edges.
    \item \textbf{Precision after refinement} $P_\text{trim} = |\{c \in C : d(c, S) \leq \varepsilon\}| \,/\, |C|$: fraction of refined Canny points that lie on the GT silhouette.
    \item \textbf{$F_{0.5}$ score}:
    \begin{equation}
    F_{0.5} = \frac{1.25 \cdot T_R \cdot P_\text{trim}}{0.25 \cdot P_\text{trim} + T_R},
    \end{equation}
    which weights precision more than recall, favouring clean contours over high coverage.
\end{itemize}
The configuration maximising $F_{0.5}$ at $\varepsilon = 2$~px is retained for each object.

\subsection{Limitations of the IoU Score for Alignment Validation }

\begin{figure}[ht]
    \centering
    \includegraphics[width=1.0\textwidth]{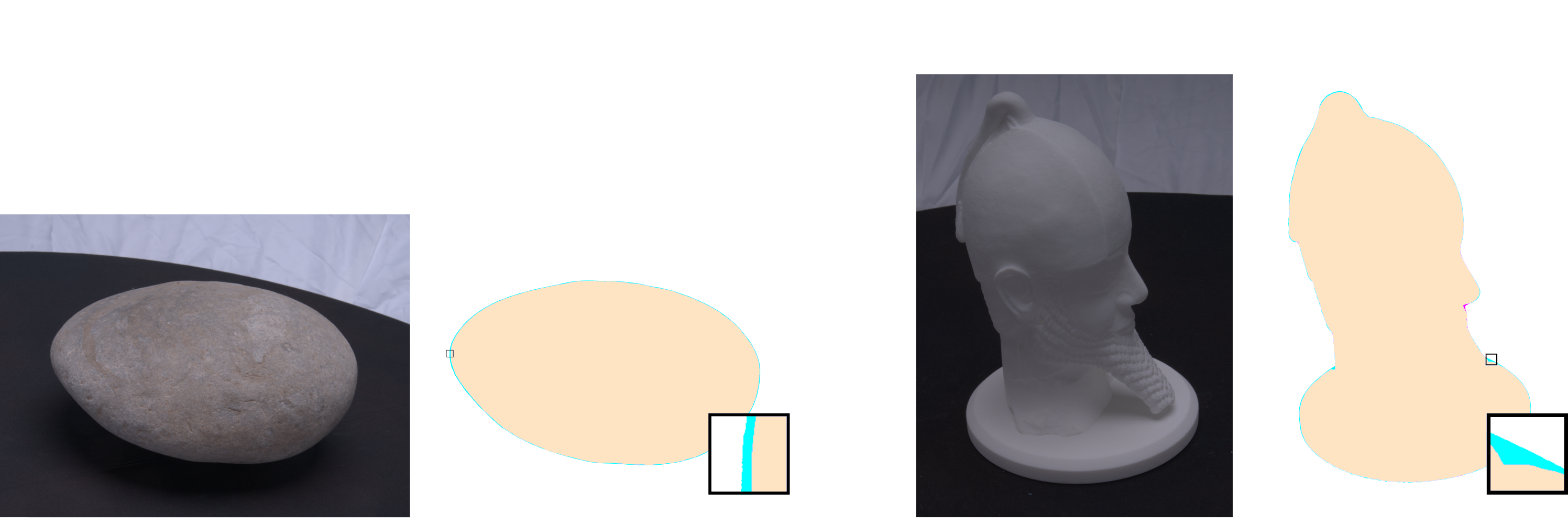}
        
    \caption{IoU limitations in assessing boundary precision. Visualisation of the discrepancy between SAM (magenta) and ground-truth (cyan) masks, with their intersection in orange. The boundary analysis reveals systematic contour shifts (4.80 and 4.51 px on average) that are critical for 2D--3D registration accuracy.}
    \label{fig:iou}
\end{figure}

EvalMVX~\cite{evalmvx} uses IoU between projected ground-truth masks and segmented image masks as its alignment-quality metric. Our analysis suggests that such global overlap scores can remain very high despite visible contour errors. On two objects from our dataset, \obj{rock} and \obj{warrior head}, SAM masks reach IoU values above 0.99 against manually segmented reference masks. However, their contours differ by 4.80 and 4.51 pixels on average, respectively, boundary shifts that are poorly captured by IoU. See Figure~\ref{fig:iou}.

As shown in Table~\ref{tab:iou}, incorporating Canny edge detection into our segmentation pipeline significantly enhances contour extraction compared to standard SAM masks. This refinement reduces the average contour distance from the 4–5 pixel range to between 1 and 2 pixels.

\begin{table}[ht!]
\centering
\caption{\textbf{Contour refinement.} The proposed Canny-based refinement significantly reduces the mean contour distance (px).}
\label{tab:alignment_comparison}
\small
\vspace{2mm}
\begin{tabular}{lccccc}
\toprule
 & \multicolumn{2}{c}{\textbf{SAM}} & & \multicolumn{2}{c}{\textbf{SAM + Canny (Ours)}} \\
\cmidrule(lr){2-3} \cmidrule(lr){5-6}
\textbf{Object} & \textbf{IoU} $\uparrow$ & \textbf{Dist. (px)} $\downarrow$ & & \textbf{IoU} $\uparrow$ & \textbf{Dist. (px)} $\downarrow$ \\
\midrule
\obj{rock}    & $0.9918$ & 4.80 & & $0.9972$ & 1.28 \\
\obj{warrior head} & $0.9913$ & 4.51 & & $0.9984$ & 1.37 \\
\bottomrule
\end{tabular}
\label{tab:iou}
\end{table}

\subsection{OpenSubstance Dataset Limitations}

\begin{figure}[ht]
    \centering
    \includegraphics[width=1.0\linewidth]{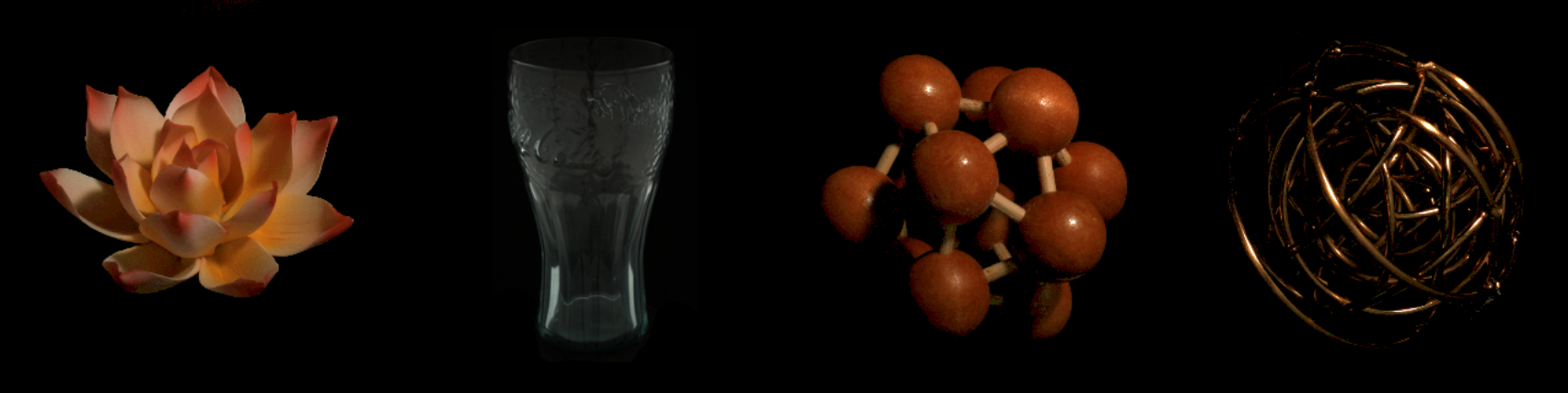}
    \caption{Sample of OpenSubstance objects that are unsuitable for rigorous 3D reconstruction analysis.}
    \label{fig:opensubs_objs}
\end{figure}

While the OpenSubstance~\cite{opensubstance} dataset provides a vast collection of objects, a significant portion of the library is unsuitable for rigorous qualitative or quantitative 3D reconstruction analysis. Figure~\ref{fig:opensubs_objs} illustrates a representative subset of these problematic cases. The first two examples possess ground-truth models, yet their geometric characteristics pose significant challenges for COLMAP; this leads to substantial registration errors between the estimated camera poses and the ground-truth geometry. The two remaining examples have highly complex geometries that could not be captured reliably by the scanning process and therefore do not have ground-truth data. This information was shared by the dataset authors.

\section{Evaluation of the View-Invariant Lighting Setup}
\label{sec:eval_inv_light}

To validate the efficacy of our single-light acquisition regime, we conducted a two-stage experimental analysis focusing on the temporal stability and the rotational invariance of the single-light setup. As detailed in the main manuscript, this setup utilises a top-mounted softbox flash and a reflecting cone encircling the turntable. When perfectly aligned, this configuration is designed to produce a view-invariant lighting environment optimised for multi-view reconstruction.  

\textbf{Photometric stability. }We first assessed the repeatability of the flash intensity by capturing 50 successive images of a static object from a fixed viewpoint. By computing the standard deviation for each pixel across the sequence, we observed an average standard deviation of 0.005 on a normalised intensity scale of $[0, 1]$. This low variance confirms the photometric stability of the flash and its suitability for consistent data acquisition.  

\begin{figure}[ht]
    \centering
    \begin{subfigure}[b]{0.30\textwidth}
        \centering
        \includegraphics[width=\textwidth]{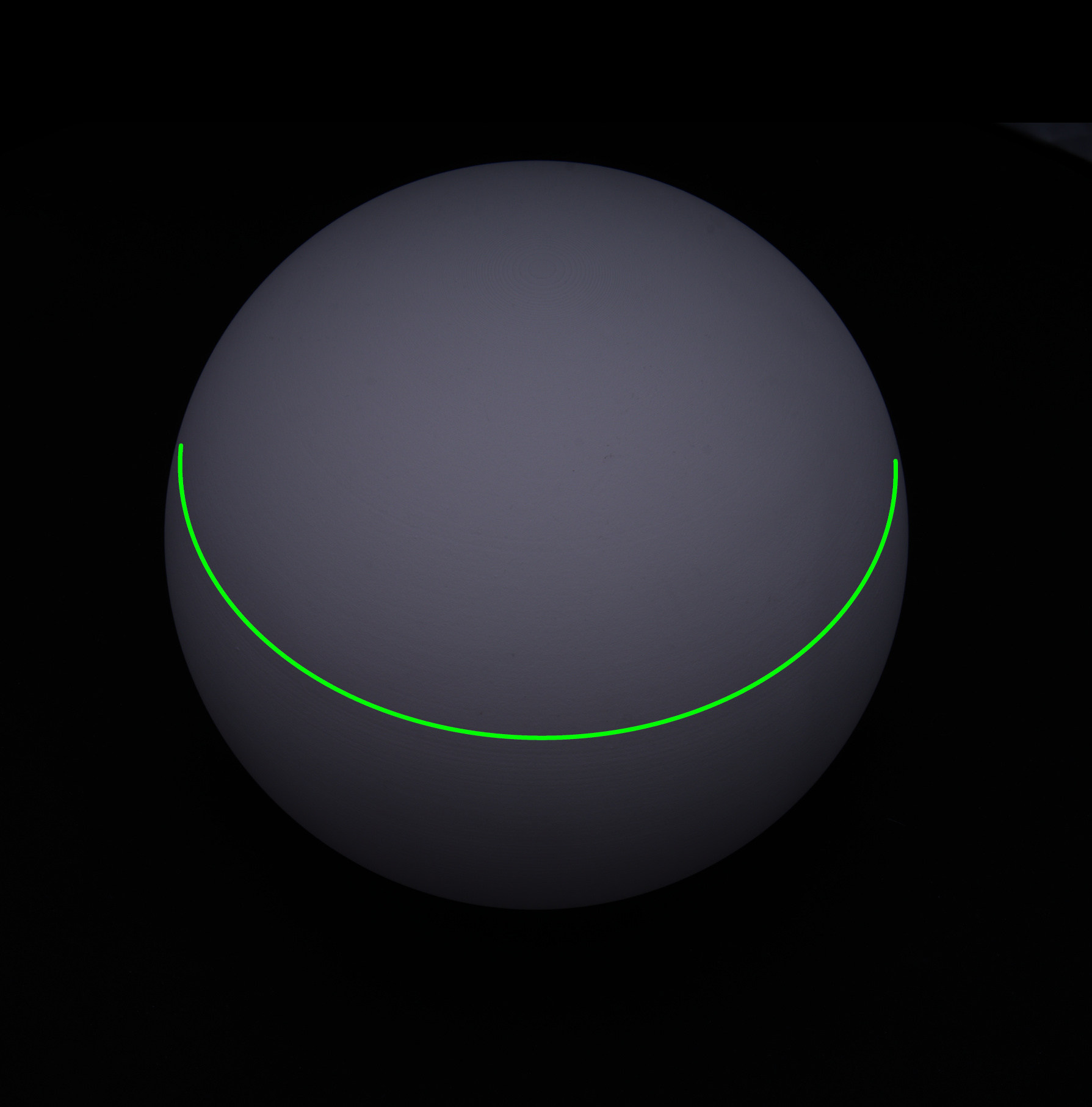}
    \end{subfigure}
    \hfill
    \begin{subfigure}[b]{0.65\textwidth}
        \centering
        \includegraphics[width=\textwidth]{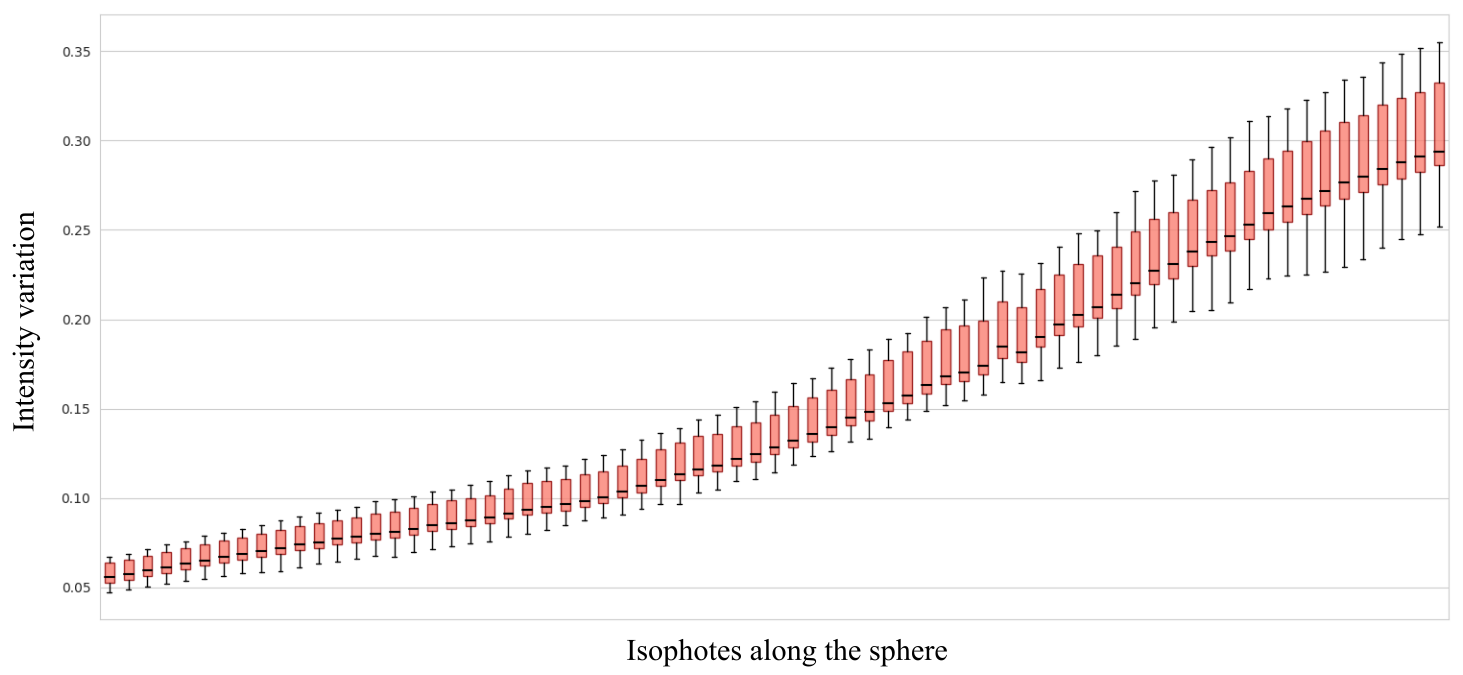}
    \end{subfigure}
    \hfill
    
     \caption{(Left) A visualisation of a single isophote (green) on the calibration sphere. (Right) Distribution of pixel intensity variations for each isophote across the sphere's height. The standard deviation ranges from 0.0054 to 0.0267, with an average of 0.0141, confirming consistent illumination across viewpoints.}
    \label{fig:rotational_invariance}
\end{figure}

\textbf{Rotational invariance. }To evaluate the view-invariant nature of the setup, we performed a symmetry analysis using a high-fidelity white sphere (25 cm in diameter) positioned at the centre of the turntable. Following camera calibration and the geometric alignment of a 3D spherical model with the image data, we identified a series of isophotes—contours of equal elevation—across the sphere’s surface. Given the rotational symmetry of the capture environment, the radiance values along these isophotes should ideally remain constant regardless of the turntable's rotation.  
We extracted intensity data from 41 viewpoints across multiple isophotes spanning the height of the sphere. The resulting distribution, visualised in the box-and-whisker plots in Figure~\ref{fig:rotational_invariance}, demonstrates the precision of our lighting invariance. The observed standard deviations range between 0.0054 and 0.0267 (mean 0.0141). Considering that a portion of this variance is attributable to the inherent flash instability identified previously, these low values confirm that the lighting remains largely independent of rotation. 

\section{Light Calibration}

\subsection{Multi-light Calibration}

Multi-light methods are evaluated on our dataset using normals predicted by UniMS-PS~\cite{unimsps}, a neural photometric-stereo method commonly used in recent multi-light reconstruction work and not requiring calibrated light sources. However, many calibrated photometric-stereo methods require light positions, directions, or intensities as input. To support such methods, we provide additional images for light calibration. Following standard sphere-based calibration procedures, we acquire multi-view multi-light sequences of two registered spheres placed in the object coordinate frame. The first is a matte sphere of diameter 60.5 mm, which can be used to estimate photometric response and intensity-related quantities. The second is a glossy sphere of diameter 60.2 mm, whose highlights provide constraints on light directions.
Because the sphere poses are known from our image-to-geometry alignment, we used them to estimate the light parameters based on two models: an anisotropic model and an isotropic model. These models are provided alongside the data. Further details regarding the light calibration are available on the project page.

\begin{figure}[ht]
    \centering
    \begin{subfigure}[b]{0.49\textwidth}
        \centering
        \includegraphics[width=\textwidth]{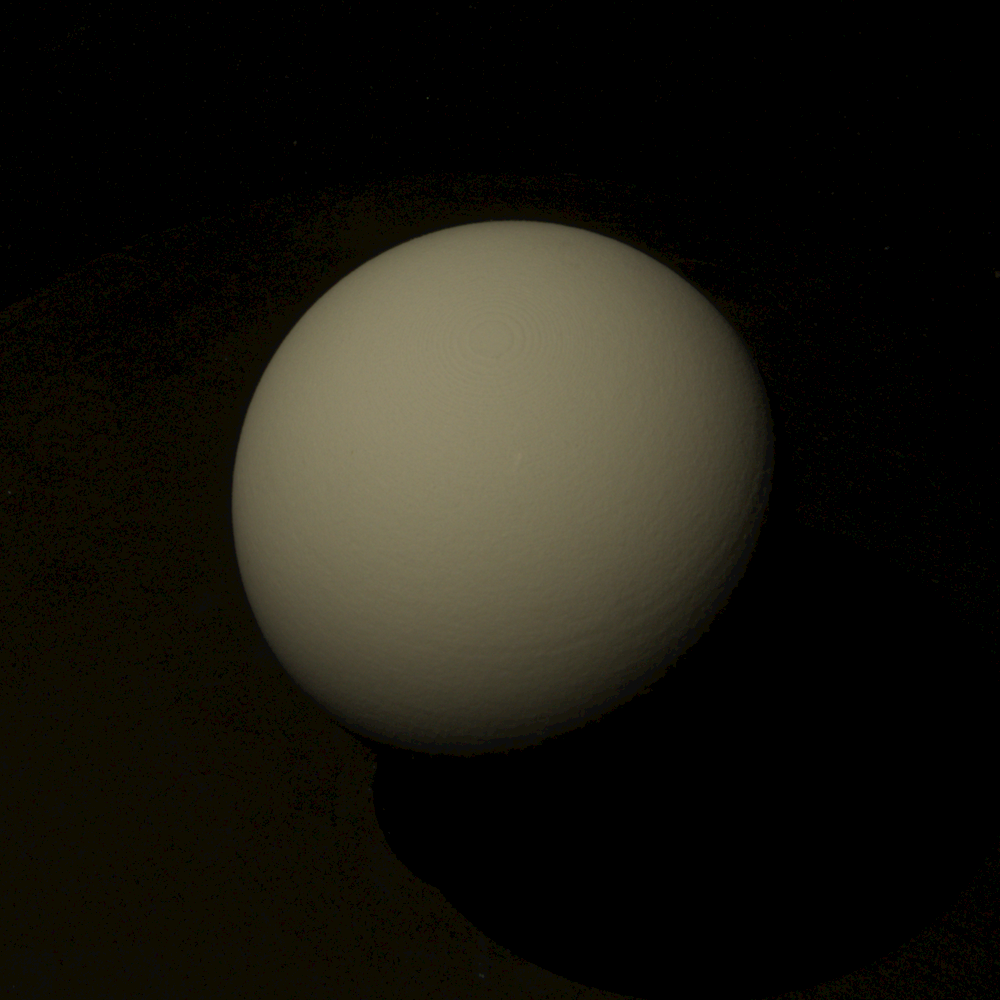}
        \label{fig:calib_sphere1}
    \end{subfigure}
    \hfill
    \begin{subfigure}[b]{0.49\textwidth}
        \centering
        \includegraphics[width=\textwidth]{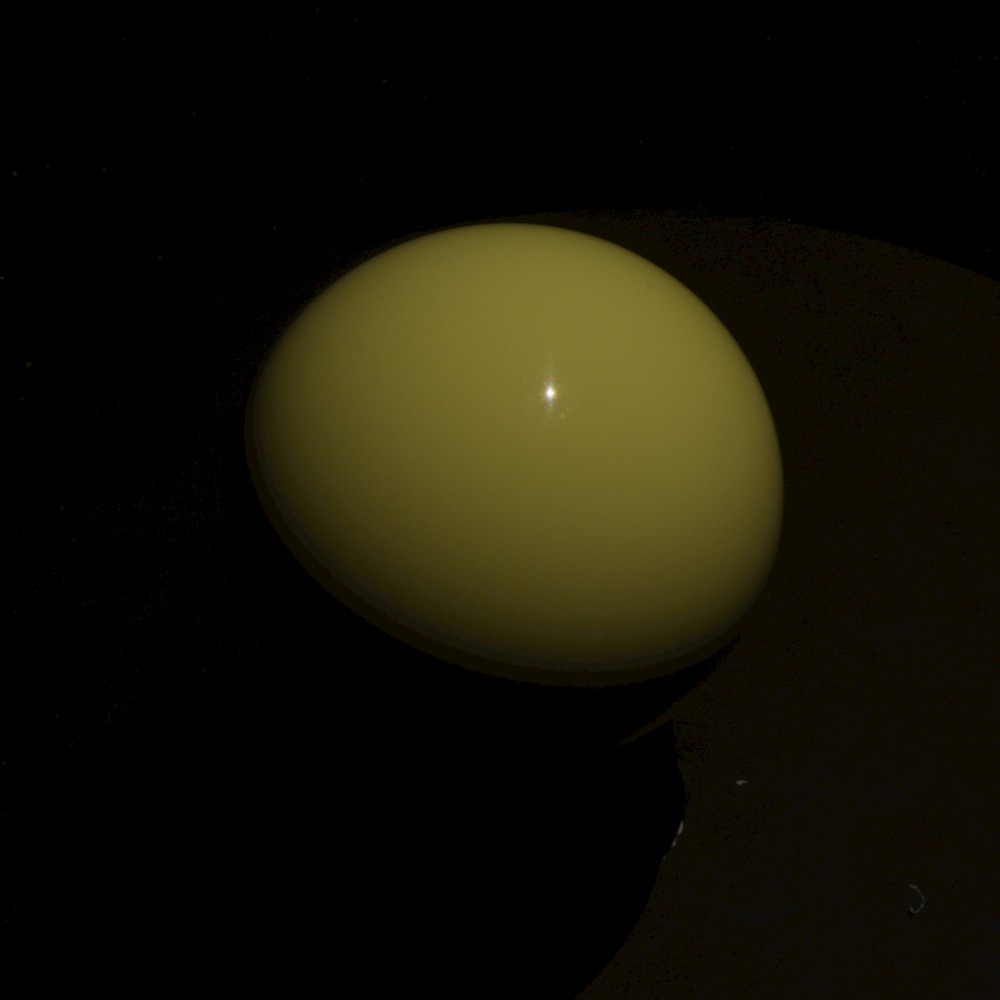}
        \label{fig:calib_sphere2}
    \end{subfigure}
    \hfill
    
    \caption{Light-calibration spheres. (Left) Matte sphere. (Right) Glossy sphere.}
    \label{fig:calib_sphere}
\end{figure}

\subsection{Photometric Calibration}

To support photometric accuracy and texture consistency, we include captures of a professional colour calibration target (Calibrite ColorChecker Classic) for each LED and the single-light scenario. These reference images allow for the correction of light colour temperatures in photometric stereo pipelines and ensure that the final reconstructed meshes feature accurate and consistent textures.

\begin{figure}[ht]
    \centering
    \begin{subfigure}[b]{0.49\textwidth}
        \centering
        \includegraphics[width=\textwidth]{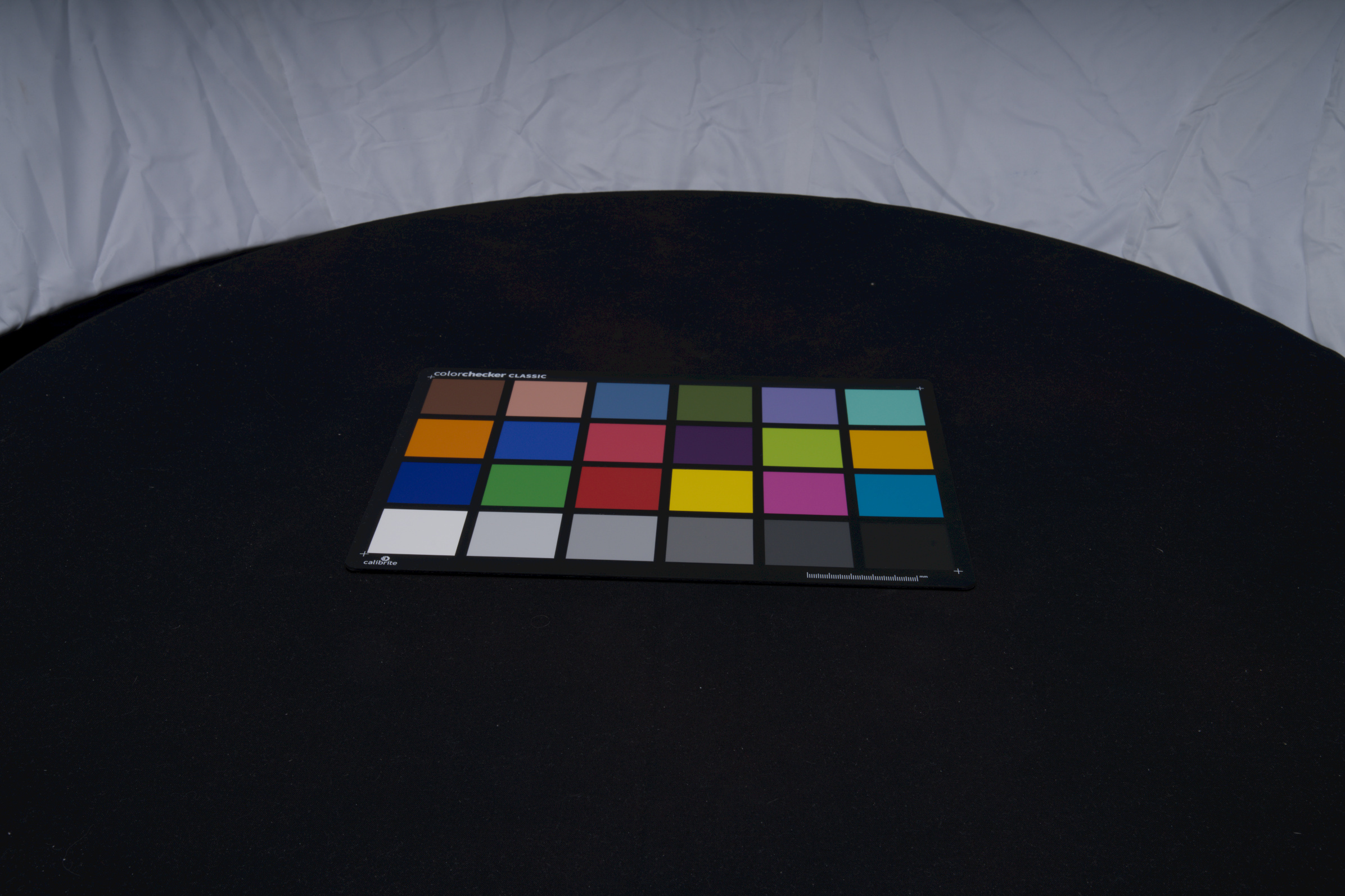}
        \label{fig:calib_color1}
    \end{subfigure}
    \hfill
    \begin{subfigure}[b]{0.49\textwidth}
        \centering
        \includegraphics[width=\textwidth]{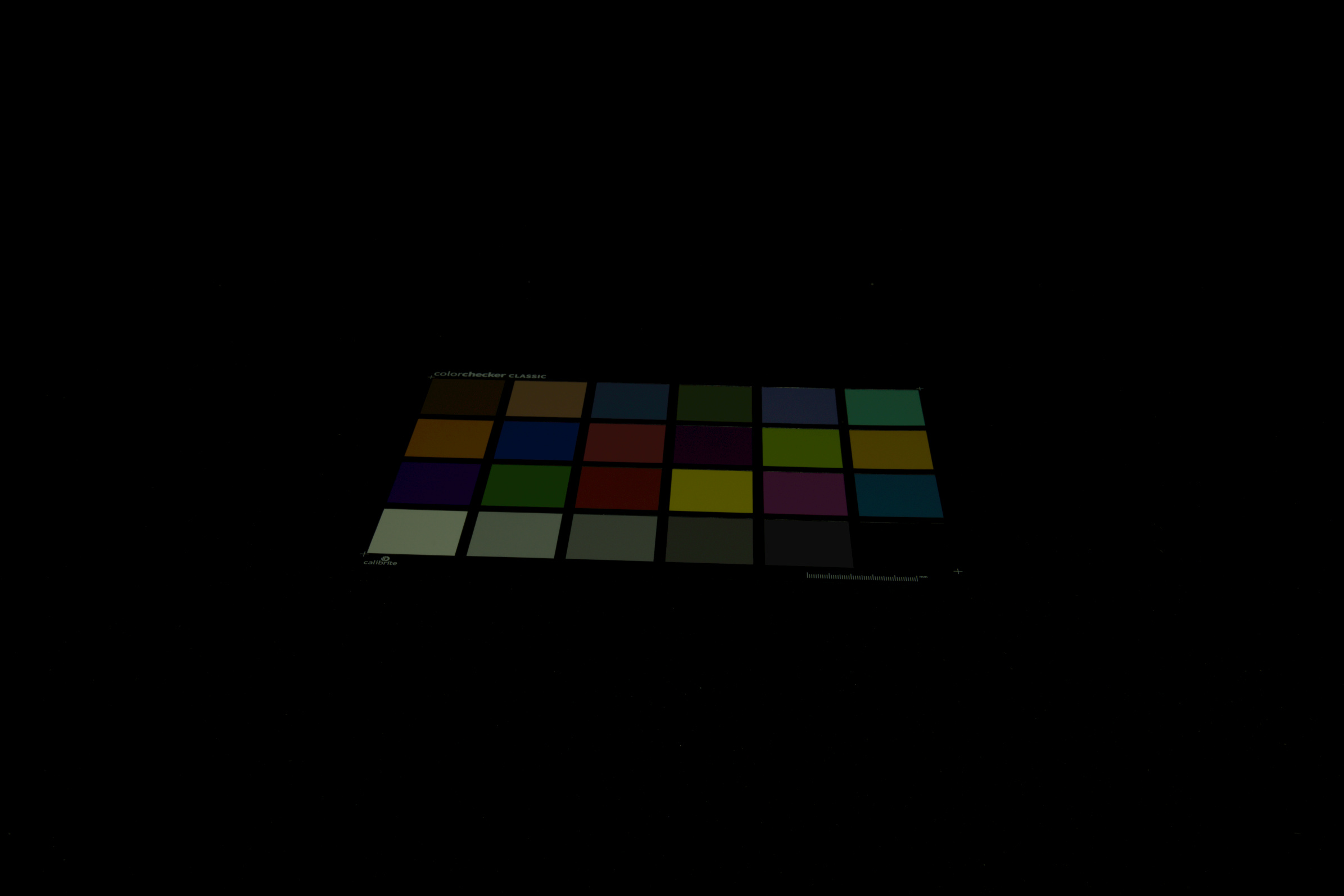}
        \label{fig:calib_color2}
    \end{subfigure}
    \hfill
    
    \caption{Colour calibration board seen in the single-light regime (left), and in the multi-light regime with one LED turned on (right).}
    \label{fig:calib_color}
\end{figure}

\section{Additional Information}

\subsection{Evaluation Protocol}
\label{sec:supp_eval}

\textbf{Mesh cleanup.} Reconstructed and ground-truth meshes are filtered in the same way, by removing vertices that are not visible from any camera view or that project outside the object masks. Mounting bases are removed from the ground-truth meshes and from the masks of the objects, and unreliable regions of the ground-truth meshes (e.g., the base of the \obj{candle} or sharp rims) are excluded using per-object 3D masks. The ground truth of \obj{wood} and \obj{sneakers} is open (no underside, no interior); reconstructed surfaces seen through these openings, in per-view masks, are removed.

\textbf{View subsets.} Subsets of $N$ views are selected by k-means clustering of the 3D camera centres ($N$ clusters, fixed seed), keeping the camera closest to each centroid; all methods using a sequence share its subsets.

All quantitative evaluations are performed by comparing the reconstructed mesh $\mathcal{P}$ against the ground-truth mesh $\mathcal{G}$. Following standard benchmarking protocols, we report precision, recall, F-score and Chamfer distance.
 
The F-score is defined as the harmonic mean of precision $P(\epsilon)$ and recall $R(\epsilon)$ for a given distance threshold $\epsilon$:
\begin{equation}
\label{eq:fscore}
F(\epsilon) = \frac{2 P(\epsilon) R(\epsilon)}{P(\epsilon) + R(\epsilon)},
\end{equation}
where precision and recall represent the fraction of points whose distance to the reference mesh is below $\epsilon$:
\begin{equation}
P(\epsilon) = \frac{1}{|\mathcal{P}|} \sum_{\hat{\mathbf{x}} \in \mathcal{P}} [d_{\hat{\mathbf{x}} \to \mathcal{G}} < \epsilon] \quad \text{and} \quad R(\epsilon) = \frac{1}{|\mathcal{G}|} \sum_{\mathbf{x} \in \mathcal{G}} [d_{\mathbf{x} \to \mathcal{P}} < \epsilon].
\end{equation}
Here, $[\cdot]$ denotes the Iverson bracket.
 
To assess the proximity between the two surfaces, we define the distance from a reconstructed point $\hat{\mathbf{x}} \in \mathcal{P}$ to the ground-truth mesh $\mathcal{G}$ as:
\begin{equation}
\label{eq:dist_p_g}
d_{\hat{\mathbf{x}} \to \mathcal{G}} = \min_{\mathbf{x} \in \mathcal{G}} \|\hat{\mathbf{x}} - \mathbf{x}\|.
\end{equation}
The distance from the ground truth to the reconstruction, $d_{\mathbf{x} \to \mathcal{P}}$, is defined analogously. These measurements are accumulated over the entire surface to compute the Chamfer distance (CD):
\begin{equation}
\label{eq:chamfer}
CD = \frac{1}{2} \left( \frac{1}{|\mathcal{P}|} \sum_{\hat{\mathbf{x}} \in \mathcal{P}} d_{\hat{\mathbf{x}} \to \mathcal{G}} + \frac{1}{|\mathcal{G}|} \sum_{\mathbf{x} \in \mathcal{G}} d_{\mathbf{x} \to \mathcal{P}} \right).
\end{equation}
Distances larger than 4~mm are discarded from the Chamfer distance; precision and recall use all points.

To ensure the reliability of these metrics, it is essential that the point densities of the reconstructed mesh $\mathcal{P}$ and the ground-truth mesh $\mathcal{G}$ are comparable. A significant discrepancy in density can bias both the Chamfer distance and the F-score, as sparsely populated regions may disproportionately affect the nearest-neighbour search. To mitigate this, we implement a consistent resampling pipeline applied to both the ground-truth and reconstructed meshes prior to evaluation. Each mesh is first converted into a dense point cloud by sampling each face on a regular grid of 0.05~mm spacing. This is followed by a spatial downsampling filter with a $0.05$ mm radius to eliminate local clusters and enforce a uniform distribution. This procedure yields, for every object, two point clouds (ground truth and reconstruction) with the same inter-point distance of about $0.05$~mm, providing a standardised and fair baseline for the final computation of the Chamfer distance and F-score.

The precision, recall and F-score curves and the radar chart pool the point distances of all objects, so that larger objects weigh more; a failed reconstruction keeps its ground-truth points in the pool, all counted as missed, which lowers recall and F-score but leaves precision unchanged.

\textbf{Curvature.} High-curvature regions are identified on the ground truth. First,
each ground-truth mesh is densified by splitting all edges longer than 0.15~mm at their midpoint, with the same threshold for every object.
The densified mesh describes exactly the same surface as the original one, but with a fine and regular sampling.
This step is needed because some ground-truth meshes, mainly those obtained by CT scanning, contain very elongated triangles near which curvature would otherwise be overestimated.

Then the curvature is computed at each vertex of the densified mesh with the Voronoi Covariance Measure (VCM)~\cite{vcm}, a robust estimator based on the shape of the Voronoi cells of the surface samples.
It depends on two radii: an offset radius $R$, which bounds the extent of each cell, and an integration radius $r$, which sets the size of the neighbourhood over which the curvature is measured.
At each vertex, it gives the two principal curvatures, of which we keep the one of largest magnitude, $k$.
Each point of the ground-truth point cloud then takes the curvature of its nearest vertex in the densified mesh.

Finally, points with $|k| > 0.75$~mm$^{-1}$ form the high-curvature region and the remaining points the flat region.
The values $R = 0.5$~mm, $r = 1$~mm and the threshold of 0.75~mm$^{-1}$ were chosen after testing several values on objects of the dataset, as the setting that best separated the high-curvature regions.
The precision, recall, F-score and Chamfer distance are computed separately on each region, which isolates the behaviour of each method on the most demanding parts of the surface.

\subsection{Data Release and Format}
\label{sec:supp_data_format}

For each object, the release provides the single-light images at four resolutions, d1 (full resolution, \(9568 \times 6376\)), d2, d4 and d8 (downsampled by factors of 2, 4 and 8 with area averaging), and the multi-light images (84 poses \(\times\) 30 LEDs) at d2, d4 and d8.
Images are stored as 16-bit half-float OpenEXR files, together with binary PNG object masks (at each lower resolution, the area average of the full-resolution mask thresholded at 128 of 255) and an AliceVision \texttt{sfm.json} file containing intrinsics and poses, with the centre of the top-left pixel at \((0,0)\).
For each evaluated object, the ground-truth kit contains the mesh, the point cloud sampled for evaluation, per-point curvature and visibility, the 3D challenge masks and the evaluation masks; the light calibration is shared by all objects.
The release totals 680\,GB and is described by a Croissant metadata file.
Converters to the input formats of all evaluated methods, with their pixel-centre conventions, and download scripts are provided in the code repository.

\subsection{Hardware and Acquisition Parameters}

The single-light acquisition setup was controlled using a Godox DP1000III-V strobe with a maximum power of 1000~W, paired with a SmallRig RA-D120 softbox to ensure uniform light diffusion. All images were captured using a Sony $\alpha$7R V camera. To maintain consistent depth of field across varying object geometries, a narrow aperture of $f/20$ and a sensitivity of ISO 100 were utilised for all captures. 

The exposure settings were adjusted according to the acquisition paradigm as follows:
\begin{itemize}
    \item \textbf{Single-light acquisition:} To effectively freeze the flash output and eliminate ambient light interference, the shutter speed was set to 1/250~s.
    \item \textbf{Multi-light acquisition:} To accommodate the LED intensities, a longer shutter speed of 1~s was employed.
\end{itemize}

\begin{figure}[ht]
    \centering
    \includegraphics[width=1.0\linewidth]{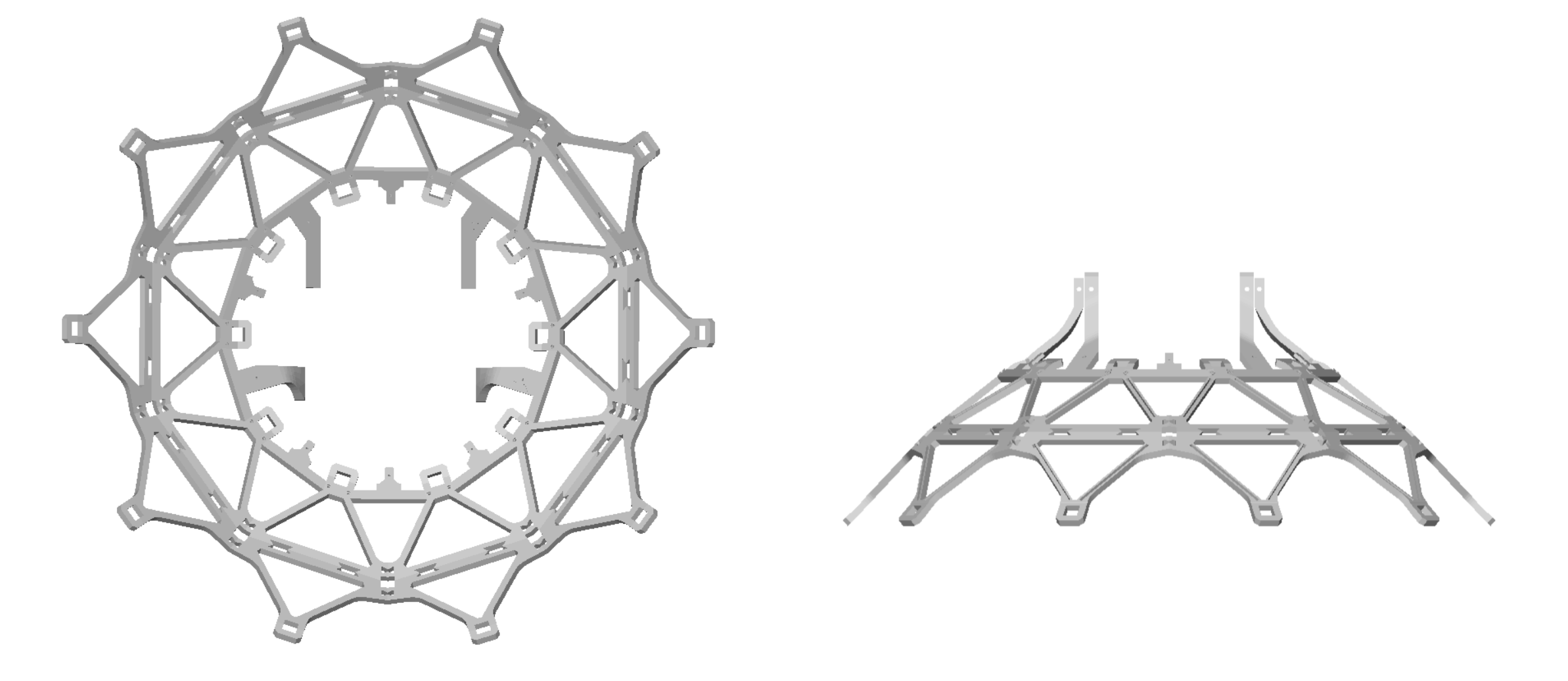}
    \caption{LED rig. Two views showing the spatial distribution of the LEDs around the camera.}
    \label{fig:led}
\end{figure}

\subsection{Baseline Settings}
\label{sec:supp_baselines}
Table~\ref{tab:baseline_versions} lists the evaluated methods and their main settings are given below; their exact versions are listed in the converters documentation of the dataset repository (footnote in Section~\ref{sec:intro}).

\begin{table}[ht!]
\centering
\caption{\textbf{Evaluated methods.}}
\label{tab:baseline_versions}
\small
\vspace{2mm}
\begin{tabular}{ll}
\toprule
\textbf{Method} & \textbf{Family} \\
\midrule
Meshroom~\cite{meshroom} & Photogrammetry \\
RealityScan~\cite{realitycapture} & Photogrammetry \\
COLMAP~\cite{colmap,colmap_mvs} & Classical MVS \\
PGSR~\cite{pgsr} & 3D Gaussian Splatting \\
NeuS2~\cite{neus2} & Neural implicit \\
Neuralangelo~\cite{na} & Neural implicit \\
SuperNormal~\cite{supernormal} & MVPS \\
RNb-NeuS2~\cite{rnbneus2} & MVPS \\
UniMS-PS~\cite{unimsps} & Photometric stereo (normals) \\
\bottomrule
\end{tabular}
\end{table}

\textbf{Common settings.}
All methods use the calibrated cameras and the images of the evaluated resolution level or view subset.
The sparse points required by COLMAP and PGSR are triangulated from these images only, with the calibrated poses kept fixed.
NeuS2, Neuralangelo, SuperNormal and RNb-NeuS2 are normalised to the minimal sphere enclosing the visual hull of the object masks, enlarged by 5\%, and their meshes are extracted by marching cubes on a $1024^3$ grid at the zero level set.
The number of iterations is adapted to the resolution level and the number of images, following the reference papers. Otherwise, each method uses the default configuration of its authors; the changes are listed below.

\textbf{Meshroom.} Release 2025.1.0.

\textbf{RealityScan.} Version 2.1.0 (Linux). Poses and calibration imported through XMP files and locked, without distortion and with the object masks. Default alignment, no image downscaling, feature detection set to high, with tie points computed from the images of the run only, followed by a high-detail reconstruction.

\textbf{COLMAP.} Version 3.12.6, default dense pipeline, with the stereo fusion restricted to the object masks.

\textbf{PGSR.} Our patched version uses the calibrated principal point, which the official code replaces by the image centre. The TSDF parameters are set to the scale of the objects (voxel 0.5\,mm, truncation 2\,mm).

\textbf{NeuS2.} Official DTU configuration.

\textbf{Neuralangelo.} Default configuration (500k iterations), at quarter resolution.

\textbf{SuperNormal.} Official configuration.

\textbf{RNb-NeuS2.} Default configuration.

\textbf{UniMS-PS.} Uncalibrated model, one inference per view on all 30 single-LED images at the evaluated resolution level, cropped to the bounding box of the object mask dilated by 5 pixels.

\subsection{Additional Results}

Table~\ref{tab:chamfer} reports the Chamfer distance and the F-score at 0.5~mm of every method on the 26 objects with ground truth, across all tested image resolutions, from d1 (full resolution) to d8 (one-eighth resolution).

\input{tables/all_cd}

Table~\ref{tab:coverage} gives, for the representative method of each family, the number of objects reconstructed for each subset of input views and each image resolution, and Table~\ref{tab:failures} lists the failed reconstructions.
These failures are real failures of the methods, not missing computations: COLMAP, PGSR and RealityScan start from a sparse model computed from the images of the evaluated configuration only, with a single attempt and default settings, and fail when this model is too sparse.

\input{tables/acquisition_coverage}
\input{tables/failures}

Across objects, view sparsity affects the methods as on the \obj{vase} (Figure~\ref{fig:resolution_views}): with 10 views, COLMAP, PGSR and RealityScan reconstruct 14, 13 and 24 of the 26 objects, and 4, 3 and 13 with 5 views, whereas NeuS2 and RNb-NeuS2 reconstruct all of them in every configuration.
At one-eighth resolution, COLMAP, PGSR and RealityScan fail on 5, 4 and 3 objects, respectively.

%% file: tables/per_corner_residuals.tex
\begin{table}[ht]
  \centering
  \caption{Distribution of the per-corner reprojection residual on the test
  checkerboard on the 20 frontal views, in pixels at the full $9568\times6376$ resolution.}
  \label{tab:per_corner_residuals}
  \small
  \begin{tabular}{lrrrrrrr}
    \toprule
    Calibration & \#corners & Mean & Median & RMS & IQR & P95 & Max \\
    \midrule
    \textbf{Textured cloth (ours)} & 6\,480 & 0.542 & 0.524 & 0.584 & 0.298 & 0.931 & 1.489 \\
    ArUco cylinder~\cite{opensubstance} & 6\,480 & 0.666 & 0.542 & 0.824 & 0.555 & 1.621 & 4.010 \\
    COLMAP  & 6\,480 & 0.809 & 0.799 & 0.913 & 0.612 & 1.559 & 2.917 \\
    ChArUco board~\cite{evalmvx} & 6\,480 & 1.086 & 0.924 & 1.312 & 0.865 & 2.541 & 5.199 \\
    \bottomrule
  \end{tabular}
\end{table}

%% file: tables/all_cd.tex
\begin{table}[htbp]
\centering
\footnotesize
\caption{Chamfer distance (mm, Eq.~\ref{eq:chamfer}) / F-score at 0.5~mm (\%) per object, method and image resolution (d1 to d8: full to one-eighth resolution), with all 84 views; Meshroom at d4 only; methods in two sub-tables, one under the other. \textbf{Bold}: best per row over both sub-tables, \underline{underline}: second best, for each metric separately (lowest Chamfer distance, highest F-score). --: failure of the method (no reconstruction, Table~\ref{tab:failures}). \textit{Mean}: point distances of all objects pooled, as if they formed a single mesh, so that larger objects weigh more; a failed reconstruction keeps its ground-truth points in the pool, all missed: they lower the F-score, and lying beyond the cutoff of Eq.~\ref{eq:chamfer} they do not enter the Chamfer distance, so Chamfer means over different numbers of reconstructed objects (\textit{Reconstructed} row) are not directly comparable.}

\resizebox{\textwidth}{!}{%
\begin{tabular}{@{}c@{}} %

\begin{tabular}{lrrrrcrrrrrr}
\toprule
 & \multicolumn{4}{c}{RealityScan} & \multicolumn{1}{c}{Meshroom} & \multicolumn{3}{c}{COLMAP} & \multicolumn{3}{c}{PGSR} \\
\cmidrule(lr){2-5} \cmidrule(lr){6-6} \cmidrule(lr){7-9} \cmidrule(lr){10-12}
Object & \multicolumn{1}{c}{d1} & \multicolumn{1}{c}{d2} & \multicolumn{1}{c}{d4} & \multicolumn{1}{c}{d8} & \multicolumn{1}{c}{d4} & \multicolumn{1}{c}{d2} & \multicolumn{1}{c}{d4} & \multicolumn{1}{c}{d8} & \multicolumn{1}{c}{d2} & \multicolumn{1}{c}{d4} & \multicolumn{1}{c}{d8} \\
\midrule
\obj{rock} & 0.108 / 99.1 & 0.106 / 99.1 & 0.148 / 98.4 & 0.261 / 90.4 & 0.108 / 98.2 & 0.102 / 98.2 & 0.115 / 97.5 & 0.156 / 94.8 & 0.340 / 87.4 & 0.508 / 68.7 & 0.774 / 12.7 \\
\obj{vase} & \textbf{0.135} / \textbf{95.9} & 0.256 / 77.4 & 0.638 / 41.3 & 1.135 / 19.1 & 0.309 / \underline{82.5} & \underline{0.243} / 79.3 & 0.267 / 73.7 & -- & 0.577 / 50.0 & 0.625 / 49.0 & -- \\
\obj{hammer head} & 0.088 / 98.9 & 0.104 / 99.0 & 0.174 / 97.8 & 0.459 / 65.3 & 0.102 / 98.4 & \underline{0.065} / 99.2 & 0.110 / 97.5 & 0.865 / 8.0 & 0.161 / 98.1 & 0.282 / 95.6 & 0.566 / 29.9 \\
\obj{candle} & 0.821 / 43.6 & 1.106 / 33.3 & 1.362 / 21.8 & 2.004 / 7.2 & 1.546 / 16.5 & 0.755 / 39.2 & 0.895 / 25.5 & -- & 0.486 / \underline{87.8} & 0.587 / 62.7 & -- \\
\obj{can} & \textbf{0.149} / \textbf{96.9} & \underline{0.184} / \underline{96.2} & 0.276 / 90.1 & 0.672 / 44.1 & 0.364 / 83.0 & 0.250 / 85.7 & 0.254 / 89.1 & 0.639 / 43.9 & 0.500 / 67.0 & 0.585 / 57.9 & 0.726 / 42.9 \\
\obj{glass} & 1.525 / 20.7 & 1.456 / 17.7 & 1.759 / 7.9 & 1.999 / 4.7 & 2.535 / 0.3 & 1.224 / 24.8 & 1.345 / 21.8 & 1.415 / 12.9 & 1.289 / 26.6 & 1.287 / 25.6 & 1.356 / 20.5 \\
\obj{petanque ball} & 0.158 / 99.0 & 0.210 / 97.2 & 0.378 / 79.2 & 0.993 / 9.1 & 0.232 / 94.4 & 0.145 / 97.5 & 0.361 / 63.9 & 1.422 / 0.2 & 0.294 / 97.1 & 0.375 / 81.3 & 0.599 / 24.5 \\
\obj{jam pot} & 0.964 / 39.8 & 1.018 / 36.8 & 1.141 / 33.1 & 1.373 / 25.1 & 1.551 / 19.2 & 0.973 / 33.9 & 0.910 / 40.0 & 1.021 / 30.9 & 0.973 / 21.4 & 1.254 / 12.8 & 1.669 / 4.7 \\
\obj{colander} & 0.408 / 69.8 & 0.415 / 70.0 & 0.731 / 38.0 & 0.929 / 30.5 & 0.460 / 65.1 & 0.622 / 54.5 & 0.551 / 61.5 & 0.620 / 53.2 & 0.394 / \underline{73.2} & \textbf{0.368} / \textbf{75.8} & \underline{0.387} / 72.5 \\
\obj{wood} & \textbf{0.121} / \underline{97.4} & 0.166 / 94.9 & 0.277 / 85.6 & 0.463 / 67.7 & 0.227 / 89.4 & 0.171 / 92.8 & 0.243 / 88.7 & 0.359 / 79.6 & 0.302 / 83.7 & 0.318 / 83.2 & 0.401 / 73.4 \\
\obj{plate} & 0.080 / 99.2 & \underline{0.077} / \textbf{99.4} & \textbf{0.069} / 99.3 & 0.149 / 95.6 & 0.088 / \underline{99.4} & 0.104 / 98.3 & 0.114 / 98.0 & 0.147 / 92.5 & 0.148 / 92.2 & 0.340 / 80.2 & 0.648 / 67.8 \\
\obj{paw pad} & \underline{0.316} / \underline{81.5} & 0.514 / 68.7 & 1.384 / 19.2 & -- & 1.928 / 3.9 & 0.991 / 3.8 & 0.886 / 12.0 & -- & 0.824 / 48.8 & 0.905 / 42.9 & 1.231 / 16.3 \\
\obj{moka pot} & \textbf{0.127} / \textbf{95.8} & \underline{0.158} / \underline{93.2} & 0.258 / 79.0 & 0.504 / 58.4 & 0.231 / 88.3 & 0.272 / 84.7 & 0.327 / 80.5 & 0.590 / 60.4 & 0.749 / 47.4 & 0.779 / 44.5 & 0.953 / 17.4 \\
\obj{lego} & 0.463 / 67.1 & 0.428 / 67.9 & 0.407 / 68.1 & 0.496 / 60.3 & 0.479 / 64.1 & 0.499 / 67.0 & 0.452 / 69.4 & 0.674 / 42.4 & 0.334 / 78.0 & 0.406 / 75.7 & 0.539 / 58.1 \\
\obj{surgery scissors} & 0.318 / 84.3 & 0.473 / 60.2 & 1.000 / 22.6 & -- & -- & 0.270 / 88.9 & 0.502 / 55.1 & -- & 0.163 / 92.5 & 0.209 / 92.6 & -- \\
\obj{knife} & 0.131 / \textbf{99.5} & 0.165 / 97.9 & 0.262 / 92.2 & 0.562 / 50.6 & 0.142 / 97.2 & \textbf{0.089} / \underline{99.2} & \underline{0.112} / 98.5 & 0.508 / 64.1 & 0.204 / 97.4 & 0.313 / 90.4 & 0.603 / 25.5 \\
\obj{spikes ball} & 0.385 / 70.2 & 0.495 / 59.8 & 0.725 / 45.4 & 0.919 / 33.6 & 0.719 / 46.9 & 0.464 / 66.0 & 0.543 / 54.4 & 1.029 / 7.4 & 0.553 / 57.5 & 0.483 / 61.4 & 0.577 / 42.5 \\
\obj{straw hat} & 0.178 / \textbf{99.4} & 0.199 / 98.0 & 0.245 / 91.9 & 0.330 / 76.9 & 0.201 / 96.9 & \underline{0.169} / \underline{98.5} & 0.200 / 96.4 & 0.253 / 91.0 & 0.283 / 87.1 & 0.307 / 84.5 & 0.404 / 65.5 \\
\obj{sneakers} & 0.162 / \underline{96.2} & \underline{0.160} / \textbf{97.3} & 0.223 / 92.9 & 0.527 / 59.7 & 0.276 / 89.4 & 0.173 / 94.6 & 0.195 / 93.6 & 0.286 / 86.1 & 0.312 / 81.2 & 0.315 / 82.1 & 0.460 / 57.5 \\
\obj{surgery saw} & \textbf{0.204} / \underline{91.4} & 0.245 / 88.4 & 0.632 / 61.1 & 1.221 / 24.6 & 0.792 / 53.1 & 0.398 / 75.9 & 0.510 / 69.2 & 1.024 / 39.6 & 0.361 / 80.5 & 0.422 / 74.0 & 0.624 / 26.8 \\
\obj{whisk} & 0.301 / 88.9 & 0.468 / 50.9 & 1.433 / 12.0 & -- & 1.488 / 14.7 & 0.397 / 66.1 & 0.598 / 33.0 & -- & 0.211 / 91.3 & 0.255 / 89.8 & -- \\
\obj{warrior head} & \underline{0.188} / \textbf{94.3} & 0.246 / 88.9 & 0.386 / 73.9 & 0.628 / 48.7 & 0.328 / 83.0 & 0.338 / 78.6 & 0.308 / 83.5 & 0.382 / 76.1 & 0.488 / 69.6 & 0.515 / 68.3 & 0.670 / 34.3 \\
\obj{plant} & \underline{0.403} / \textbf{78.7} & 0.579 / 59.5 & 0.690 / 47.2 & 0.982 / 29.3 & 0.664 / 56.9 & 0.657 / 52.3 & 0.736 / 44.8 & 0.970 / 27.6 & 0.619 / 50.9 & 0.622 / 52.2 & 0.712 / 31.8 \\
\obj{grid} & \textbf{0.216} / \underline{93.4} & 0.363 / 81.9 & 0.428 / 60.9 & 0.838 / 28.1 & 0.408 / 64.1 & 0.434 / 74.2 & 0.601 / 63.2 & 0.875 / 27.1 & 0.347 / 78.2 & 0.346 / 75.6 & 0.568 / 49.6 \\
\obj{bowl} & \textbf{0.276} / \textbf{90.5} & 0.484 / 73.4 & 1.024 / 34.2 & 1.791 / 6.6 & 1.022 / 37.9 & \underline{0.364} / \underline{79.1} & 0.729 / 45.7 & 0.927 / 15.8 & 1.134 / 24.5 & 1.175 / 20.3 & 1.190 / 14.8 \\
\obj{spine} & 0.465 / 70.6 & 0.528 / 65.1 & 0.732 / 43.6 & 1.022 / 26.7 & 0.742 / 46.1 & 0.508 / 65.5 & 0.579 / 58.4 & 0.843 / 33.4 & 0.569 / 61.7 & 0.552 / 65.3 & 0.632 / 43.3 \\
\midrule
\textit{Mean} & 0.333 / 82.4 & 0.381 / 77.4 & 0.536 / 63.6 & 0.731 / 45.4 & 0.477 / 68.1 & 0.380 / 74.5 & 0.432 / 70.6 & 0.557 / 54.7 & 0.503 / 65.9 & 0.559 / 62.1 & 0.702 / 40.1 \\
\textit{Reconstructed} & 26/26 & 26/26 & 26/26 & 23/26 & 25/26 & 26/26 & 26/26 & 21/26 & 26/26 & 26/26 & 22/26 \\
\bottomrule
\end{tabular} \\ 

\noalign{\vspace{0.4cm}} %

\begin{tabular}{lrrrrrrrr}
\toprule
 & \multicolumn{3}{c}{NeuS2} & Neuralangelo & \multicolumn{3}{c}{RNb-NeuS2} & SuperNormal \\
\cmidrule(lr){2-4} \cmidrule(lr){5-5} \cmidrule(lr){6-8} \cmidrule(lr){9-9}
Object & \multicolumn{1}{c}{d2} & \multicolumn{1}{c}{d4} & \multicolumn{1}{c}{d8} & \multicolumn{1}{c}{d4} & \multicolumn{1}{c}{d2} & \multicolumn{1}{c}{d4} & \multicolumn{1}{c}{d8} & \multicolumn{1}{c}{d4} \\
\midrule
\obj{rock} & 0.103 / 99.6 & 0.085 / 99.8 & 0.098 / 99.8 & 0.099 / 98.4 & \underline{0.068} / \textbf{100.0} & 0.108 / \underline{99.9} & 0.370 / 82.1 & \textbf{0.063} / 99.8 \\
\obj{vase} & 0.271 / 67.0 & 0.268 / 66.9 & 0.302 / 65.4 & 0.276 / 81.8 & 0.610 / 60.7 & 0.495 / 67.7 & 0.529 / 68.4 & 0.947 / 42.9 \\
\obj{hammer head} & 0.105 / 99.1 & 0.096 / \underline{99.8} & 0.105 / 99.7 & \textbf{0.058} / \textbf{99.8} & 0.078 / 99.2 & 0.081 / 99.2 & 0.160 / 97.2 & 0.147 / 97.5 \\
\obj{candle} & \underline{0.253} / 86.6 & 0.257 / 86.0 & 0.263 / 86.3 & \textbf{0.247} / \textbf{89.9} & 0.304 / 81.0 & 0.292 / 81.8 & 0.365 / 75.1 & 1.482 / 0.8 \\
\obj{can} & 0.420 / 78.8 & 0.388 / 79.9 & 0.412 / 79.7 & 0.307 / 83.6 & 0.515 / 68.0 & 0.528 / 67.1 & 0.708 / 53.1 & 0.563 / 51.7 \\
\obj{glass} & 1.051 / 31.1 & 0.880 / 46.0 & \textbf{0.834} / \underline{46.6} & \underline{0.837} / \textbf{47.8} & 1.256 / 19.5 & 1.247 / 19.3 & 1.258 / 20.1 & 2.006 / 5.1 \\
\obj{petanque ball} & 0.218 / 92.8 & 0.162 / 93.9 & 0.185 / 90.6 & 0.178 / 94.3 & 0.113 / 99.3 & \textbf{0.074} / \underline{99.6} & \underline{0.088} / \textbf{99.7} & 0.126 / 96.7 \\
\obj{jam pot} & \textbf{0.383} / \textbf{77.4} & \underline{0.386} / \underline{76.3} & 0.438 / 70.6 & 0.550 / 63.9 & 1.203 / 9.9 & 1.254 / 10.2 & 1.213 / 10.4 & 1.255 / 10.9 \\
\obj{colander} & 0.805 / 42.1 & 0.782 / 44.0 & 0.796 / 41.2 & 0.599 / 51.8 & 0.574 / 51.7 & 0.569 / 52.8 & 0.776 / 39.7 & 0.488 / 62.6 \\
\obj{wood} & 0.269 / 87.3 & 0.272 / 87.0 & 0.308 / 84.9 & 0.225 / 89.7 & 0.185 / \textbf{97.5} & 0.161 / 96.8 & 0.172 / 96.5 & \underline{0.145} / 94.1 \\
\obj{plate} & 0.106 / 98.0 & 0.133 / 96.5 & 0.154 / 96.1 & 0.138 / 95.5 & 1.793 / 9.7 & 1.650 / 9.7 & 1.635 / 10.9 & 1.783 / 10.0 \\
\obj{paw pad} & 0.461 / 71.9 & 0.461 / 71.2 & 0.460 / 69.6 & \textbf{0.257} / \textbf{85.9} & 0.874 / 33.0 & 0.881 / 33.4 & 0.906 / 31.5 & 1.327 / 18.4 \\
\obj{moka pot} & 0.485 / 68.0 & 0.415 / 73.7 & 0.357 / 79.0 & 0.354 / 74.1 & 0.451 / 71.6 & 0.300 / 81.8 & 0.268 / 84.0 & 0.486 / 63.5 \\
\obj{lego} & \underline{0.326} / \underline{80.1} & 0.332 / 79.8 & 0.334 / 79.5 & \textbf{0.294} / \textbf{81.8} & 0.604 / 59.2 & 0.539 / 61.1 & 0.613 / 58.2 & 0.723 / 56.9 \\
\obj{surgery scissors} & 0.080 / 99.3 & 0.089 / 99.0 & 0.083 / 99.1 & 1.691 / 8.5 & \textbf{0.066} / \textbf{100.0} & \underline{0.073} / \underline{100.0} & 0.086 / 99.8 & 0.084 / 99.3 \\
\obj{knife} & 0.134 / 98.4 & 0.141 / 97.7 & 0.141 / 98.4 & 0.278 / 89.3 & 0.231 / 87.5 & 0.198 / 91.2 & 0.216 / 89.7 & 0.380 / 77.1 \\
\obj{spikes ball} & 0.304 / 80.8 & 0.249 / 86.9 & 0.247 / 87.7 & 0.232 / 87.9 & 0.193 / 97.0 & \underline{0.166} / \underline{98.1} & \textbf{0.137} / \textbf{98.3} & 0.256 / 93.3 \\
\obj{straw hat} & 0.229 / 91.0 & 0.227 / 91.4 & 0.224 / 91.6 & \textbf{0.167} / 97.9 & 0.283 / 88.6 & 0.234 / 91.3 & 0.292 / 85.5 & 0.283 / 88.2 \\
\obj{sneakers} & 0.265 / 82.8 & 0.288 / 81.4 & 0.282 / 81.2 & \textbf{0.151} / 91.4 & 0.279 / 82.8 & 0.262 / 84.4 & 0.269 / 84.3 & 0.285 / 82.8 \\
\obj{surgery saw} & 0.256 / 87.6 & 0.252 / 87.6 & 0.241 / 88.9 & \underline{0.214} / \textbf{92.3} & 0.313 / 80.9 & 0.300 / 83.2 & 0.338 / 79.1 & 0.444 / 67.3 \\
\obj{whisk} & 0.126 / 96.2 & 0.124 / 97.8 & 0.134 / 95.6 & -- & \underline{0.084} / \underline{98.9} & \textbf{0.065} / \textbf{99.2} & 0.109 / 98.5 & 0.097 / 98.1 \\
\obj{warrior head} & 0.472 / 72.3 & 0.485 / 71.4 & 0.471 / 72.8 & \textbf{0.175} / \underline{94.1} & 0.277 / 86.5 & 0.283 / 86.4 & 0.284 / 86.6 & 0.373 / 78.4 \\
\obj{plant} & 0.517 / 68.8 & 0.533 / 67.4 & 0.545 / 66.9 & 0.418 / 74.9 & 0.531 / 64.5 & \textbf{0.395} / \underline{76.0} & 0.422 / 73.8 & 0.462 / 68.5 \\
\obj{grid} & \underline{0.219} / \textbf{93.5} & 0.222 / 93.1 & 0.235 / 91.6 & 1.011 / 25.1 & 0.248 / 86.6 & 0.224 / 89.2 & 0.260 / 86.1 & 0.291 / 83.8 \\
\obj{bowl} & 0.546 / 50.5 & 0.591 / 48.6 & 0.560 / 51.3 & 0.824 / 44.3 & 0.864 / 34.0 & 0.927 / 35.7 & 0.999 / 31.9 & 1.165 / 24.5 \\
\obj{spine} & 0.437 / 71.0 & 0.434 / 71.7 & 0.447 / 70.8 & 0.368 / 80.8 & \textbf{0.202} / \textbf{91.2} & \underline{0.228} / \underline{90.1} & 0.296 / 85.2 & 0.404 / 77.4 \\
\midrule
\textit{Mean} & 0.393 / 75.0 & 0.388 / 75.5 & 0.391 / 75.3 & 0.353 / 77.9 & 0.501 / 64.0 & 0.469 / 66.7 & 0.540 / 62.4 & 0.571 / 60.8 \\
\textit{Reconstructed} & 26/26 & 26/26 & 26/26 & 25/26 & 26/26 & 26/26 & 26/26 & 26/26 \\
\bottomrule
\end{tabular}

\end{tabular}%
}
\label{tab:chamfer}
\end{table}

%% file: tables/acquisition_coverage.tex
\begin{table}[htbp]
\centering
\footnotesize
\caption{Number of objects reconstructed (out of 26) per method, for each subset of input views at quarter resolution (d4) and for each image resolution with all 84 views. A missing reconstruction is a failure of the method (Table~\ref{tab:failures}). --: configuration not run; at full resolution, the methods other than RealityScan fail for lack of GPU memory (Section~\ref{sec:protocol}).}
\label{tab:coverage}
\begin{tabular}{lccccccccc}
\toprule
 & \multicolumn{5}{c}{Views (d4)} & \multicolumn{4}{c}{Resolution (84 views)} \\
\cmidrule(lr){2-6} \cmidrule(lr){7-10}
Method & 84 & 40 & 20 & 10 & 5 & d1 & d2 & d4 & d8 \\
\midrule
RealityScan & 26 & 26 & 25 & 24 & 13 & 26 & 26 & 26 & 23 \\
COLMAP & 26 & 24 & 21 & 14 & 4 & -- & 26 & 26 & 21 \\
PGSR & 26 & 24 & 21 & 13 & 3 & -- & 26 & 26 & 22 \\
NeuS2 & 26 & 26 & 26 & 26 & 26 & -- & 26 & 26 & 26 \\
RNb-NeuS2 & 26 & 26 & 26 & 26 & 26 & -- & 26 & 26 & 26 \\
\bottomrule
\end{tabular}
\end{table}

%% file: tables/failures.tex
\begin{table}[htbp]
\centering
\footnotesize
\caption{Failed reconstructions per configuration. COLMAP and PGSR fail when structure from motion on the images of the configuration alone yields no usable sparse model (single attempt, default settings, no fallback); the other failures are runs that produced no mesh. Failed objects are excluded from all aggregated scores and declared as $n/26$ in the captions.}
\label{tab:failures}
\begin{tabular}{lcp{0.62\textwidth}}
\toprule
Configuration & Failed & Objects \\
\midrule
Meshroom d4 & 1/26 & \obj{surgery scissors} \\
RealityScan d4, 20 views & 1/26 & \obj{surgery scissors} \\
RealityScan d4, 10 views & 2/26 & \obj{surgery scissors}, \obj{whisk} \\
RealityScan d4, 5 views & 13/26 & \obj{candle}, \obj{glass}, \obj{petanque ball}, \obj{moka pot}, \obj{lego}, \obj{surgery scissors}, \obj{spikes ball}, \obj{surgery saw}, \obj{whisk}, \obj{plant}, \obj{grid}, \obj{bowl}, \obj{spine} \\
RealityScan d8 & 3/26 & \obj{paw pad}, \obj{surgery scissors}, \obj{whisk} \\
COLMAP d4, 40 views & 2/26 & \obj{paw pad}, \obj{surgery scissors} \\
COLMAP d4, 20 views & 5/26 & \obj{candle}, \obj{paw pad}, \obj{surgery scissors}, \obj{spikes ball}, \obj{whisk} \\
COLMAP d4, 10 views & 12/26 & \obj{candle}, \obj{can}, \obj{glass}, \obj{petanque ball}, \obj{colander}, \obj{paw pad}, \obj{surgery scissors}, \obj{spikes ball}, \obj{surgery saw}, \obj{whisk}, \obj{bowl}, \obj{spine} \\
COLMAP d4, 5 views & 22/26 & \obj{vase}, \obj{hammer head}, \obj{candle}, \obj{can}, \obj{glass}, \obj{petanque ball}, \obj{jam pot}, \obj{colander}, \obj{paw pad}, \obj{moka pot}, \obj{lego}, \obj{surgery scissors}, \obj{knife}, \obj{spikes ball}, \obj{straw hat}, \obj{surgery saw}, \obj{whisk}, \obj{warrior head}, \obj{plant}, \obj{grid}, \obj{bowl}, \obj{spine} \\
COLMAP d8 & 5/26 & \obj{vase}, \obj{candle}, \obj{paw pad}, \obj{surgery scissors}, \obj{whisk} \\
PGSR d4, 40 views & 2/26 & \obj{paw pad}, \obj{surgery scissors} \\
PGSR d4, 20 views & 5/26 & \obj{candle}, \obj{glass}, \obj{paw pad}, \obj{surgery scissors}, \obj{whisk} \\
PGSR d4, 10 views & 13/26 & \obj{candle}, \obj{can}, \obj{glass}, \obj{petanque ball}, \obj{colander}, \obj{paw pad}, \obj{surgery scissors}, \obj{spikes ball}, \obj{surgery saw}, \obj{whisk}, \obj{grid}, \obj{bowl}, \obj{spine} \\
PGSR d4, 5 views & 23/26 & \obj{vase}, \obj{hammer head}, \obj{candle}, \obj{can}, \obj{glass}, \obj{petanque ball}, \obj{jam pot}, \obj{colander}, \obj{paw pad}, \obj{moka pot}, \obj{lego}, \obj{surgery scissors}, \obj{knife}, \obj{spikes ball}, \obj{straw hat}, \obj{sneakers}, \obj{surgery saw}, \obj{whisk}, \obj{warrior head}, \obj{plant}, \obj{grid}, \obj{bowl}, \obj{spine} \\
PGSR d8 & 4/26 & \obj{vase}, \obj{candle}, \obj{surgery scissors}, \obj{whisk} \\
Neuralangelo d4 & 1/26 & \obj{whisk} \\
\bottomrule
\end{tabular}
\end{table}